\documentclass[twoside,leqno,twocolumn]{article}
\usepackage{ltexpprt}
\usepackage{epsfig}
\usepackage{graphicx}
\usepackage{graphics}
\usepackage{amssymb}
\usepackage{boxedminipage}

\begin{document}

\title{\Large Non-Negative Matrix Factorization, Convexity and Isometry
\thanks{Supported by Google Grants}}
\author{Nikolaos Vasiloglou  \\
\and
Alexander G. Gray \\
\and
David V. Anderson\thanks{Georgia Institute of Technology}}
\date{}

\maketitle
\begin{abstract}
\small\baselineskip=9pt In this paper we explore avenues for
improving the reliability of dimensionality reduction methods such
as Non-Negative Matrix Factorization (NMF) as interpretive
exploratory data analysis tools.  We first explore the difficulties
of the optimization problem underlying NMF, showing for the first
time that non-trivial NMF solutions always exist and that the
optimization problem is actually convex, by using the theory of
Completely Positive Factorization.  We subsequently explore four
novel approaches to finding globally-optimal NMF solutions using
various ideas from convex optimization.  We then develop a new
method, isometric NMF (isoNMF), which preserves non-negativity while
also providing an isometric embedding, simultaneously achieving two
properties which are helpful for interpretation.  Though it results
in a more difficult optimization problem, we show experimentally
that the resulting method is scalable and even achieves more compact
spectra than standard NMF.
\end{abstract}

\section{Introduction.}
In this paper we explore avenues for improving the reliability of
dimensionality reduction methods such as Non-Negative Matrix
Factorization (NMF) \cite{lee1999lpo} as interpretive exploratory
data analysis tools, to make them reliable enough for, say, making
scientific conclusions from astronomical data.

NMF is a dimensionality reduction method of much recent interest
which can, for some common kinds of data, sometimes yield results
which are more meaningful than those returned by the classical
method of Principal Component Analysis (PCA), for example (though it
will not in general yield better dimensionality reduction than PCA,
as we'll illustrate later).  For data of significant interest such
as images (pixel intensities) or text (presence/absence of words) or
astronomical spectra (magnitude in various frequencies), where the
data values are non-negative, NMF can produce components which can
themselves be interpreted as objects of the same type as the data
which are added together to produce the observed data.  In other
words, the components are more likely to be sensible images or
documents or spectra.  This makes it a potentially very useful
interpretive data mining tool for such data.

A second important interpretive usage of dimensionality reduction
methods is the plot of the data points in the low-dimensional space
obtained (2-D or 3-D, generally).  Multidimensional scaling methods
and recent nonlinear manifold learning methods
focus on this usage, typically enforcing that the distances between
the points in the original high-dimensional space are preserved in
the low-dimensional space (isometry constraints).  Then, apparent
relationships in the low-D plot (indicating for example cluster
structure or outliers) correspond to actual relationships.  A plot
of the points using components found by standard NMF methods will in
general produce misleading results in this regard, as existing
methods do not enforce such a constraint.

Another major reason that NMF might not yield reliable interpretive
results is that current optimization methods \cite{kim:nnm,
lee2001ann} might not find the actual optimum, leading to poor
performance in terms of both of the above interpretive usages. This
is because its objective function is not convex, and so
unconstrained optimizers are used. Thus, obtaining a reliably
interpretable NMF method requires understanding its optimization
problem more deeply -- especially if we are going to actually create
an additionally difficult optimization problem by adding isometry
constraints.

\subsection{Paper Organization.}
In Section \ref{sec_Convexity} we first study at a fundamental level
the optimization problem of standard NMF.  We relate for the first
time the NMF problem to the theory of Completely Positive
Factorization, then using that theory, we show that every
non-negative matrix has a non-trivial exact non-negative matrix
factorization of the form W=VH, a basic fact which had not been
shown until now.  Using this theory we also show that a convex
formulation of the NMF optimization problem exists, though a
practical solution method for this formulation does not yet exist.
We then explore four novel formulations of the NMF optimization
problem toward achieving a global optimum: convex relaxation using
the positive semidefinite cone, approximating the semidefinite cone
with smaller ones, convex multi-objective optimization, and
generalized geometric programming. We highlight the difficulties
encountered by each approach.

In order to turn to the question of how to create a new isometric
NMF, in Section \ref{sec_Isometric_Embedding} we give background on
two recent successful manifold learning methods, Maximum Variance
Unfolding (MVU) \cite{weinberger2006ind} and a new variant of it,
Maximum Furthest Neighbor Unfolding (MFNU) \cite{vasiloglou2008ssm}.
It has been shown experimentally \cite{weinberger2004lkm} that they can recover the intrinsic
dimension of a dataset very reliably and effectively, compared to
previous well-known methods such as ISOMAP \cite{tenenbaum2000ggf},
Laplacian Eigen-Maps \cite{belkin2003led} and Diffusion Maps
\cite{coifman2006dm}. In synthetic experiments the above methods
manage to decompose data into meaningful dimensions. For example,
for a dataset consisting of images of a statue photographed from
different horizontal and vertical angles, MVU and MFNU find two
dimensions that can be identified as the horizontal and the vertical
camera angle.  MVU and MFNU contain ideas, particularly concerning
the formulation of their optimization problems, upon which isometric
NMF will be based.

In Section \ref{sec_Isometric_NMF} we show a practical algorithm for
an isometric NMF (\textit{isoNMF} for short), representing a new
data mining method capable of producing both interpretable
components and interpretable plots simultaneously.  We use ideas for
efficient optimization and efficient neighborhood computation to
obtain a practical scalable method.

In Section \ref{sec_Experimental_Results} we demonstrate the utility
of isoNMF in experiments with datasets used in previous papers.  We
show that the components it finds are comparable to those found by
standard NMF, while it additionally preserves distances much better,
and also results in more compact spectra.

\section{Convexity in Non Negative Matrix Factorization.}
\label{sec_Convexity}
Given a non-negative matrix $V \in \Re_{+}^{N
\times m}$ the goal of NMF is to decompose it in two matrices $W \in
\Re^{N \times k}_{+}$, $H \in \Re^{k\times m}_{+}$ such that $V=WH$.
Such a factorization always exists  for $k \geq m$. The
factorization has a trivial solution where $W=V$ and $H=I_m$.
Determining the minimum $k$ is a difficult problem and no algorithm
exists for finding it. In general we can show that NMF can be cast
as a Completely Positive (CP) Factorization problem
\cite{berman2003cpm}.
\begin{Definition}
A matrix $A \in \Re_{+}^{N\times N}$ is Completely Positive if it
can be factored in the form $A=BB^T$, where $B \in \Re_{+}^{N\times
k}$. The minimum $k$ for which $A=BB^T$ holds is called the CP rank
of $A$.
\end{Definition}
Not all matrices admit a completely positive factorization even if
they are positive definite and non-negative. Notice though that for
every positive definite non-negative matrix a Cholesky factorization
always exists, but there is no guarantee that the Cholesky factors
are non-negative too. Up to now there is no algorithm of polynomial
complexity that can decide if a given positive matrix is CP. A
simple observation can show that $A$ has to be positive definite,
but this is a necessary and not a sufficient condition.
\begin{theorem}
\label{cp_rank} If $A \in \Re_{+}^{N\times N}$ is CP then
$rank(A)\leq \mbox{cp-rank(A)} \leq \frac{N(N+1)}{2}-1$
\end{theorem}
The proof can be found in \cite{berman2003cpm}p.156. It is also
conjectured that the upper bound can be tighter $\frac{N^2}{4}$.

\begin{theorem}
\label{diagonal_dominance} if $A \in \Re_{+}^{N\times N}$
is diagonally dominant\footnote{A matrix is diagonally dominant if
$a_{ii}\geq \sum_{j \neq i} |a_ij|$}, then it is also CP.
\end{theorem}
The proof of the theorem can be found in \cite{kaykobad1987nfm}.
Although CP factorization ($A=BB^T$) doesn't exist for every matrix,
we prove that non-trivial NMF ($A=WH$) always exists.
\begin{theorem}
\label{non_triviality} Every non-negative matrix $V \in \Re_{+}^{N
\times m}$ has a non-trivial, non-negative factorization of the form
$V=WH$.
\end{theorem}

\begin{proof}
Consider the following matrix:
\begin{equation}
Z=
\left[
\begin{array}{cc}
  D & V \\
  V^T & E
\end{array}
\right]
\end{equation}
We want to prove that there always exists $B \in \Re_{+}^{N\times
k}$ such that $Z=BB^T$. If this is true then $B$ can take the form:
\begin{equation}
B=\left[
    \begin{array}{c}
      W \\
      H^T
    \end{array}
  \right]
\end{equation}
Notice that if $D$ and $E$ are arbitrary diagonally dominant
completely positive matrices, then $B$ always exists. The simplest
choice would be to chose them as diagonal matrices where each
element is greater or equal to the sum of rows/columns of V. Since
they are diagonally dominant
 according to \ref{diagonal_dominance} $Z$ is always CP. Since $Z$ is CP then $B$
exists so do $W$ and $H$. $\square$
\end{proof}
Although theorem \ref{diagonal_dominance} also provides an algorithm
for constructing  the CP-factorization, the cp-rank is usually high.
A corollary of theorems \ref{cp_rank} ($\mbox{cp-rank}(A) \geq \mbox{rank}(A)$) and \ref{non_triviality} (existence of NMF)   is
that SVD has always a more compact spectrum than NMF.

There is no algorithm known yet for computing an exact NMF despite
its existence. In practice, scientists try to minimize the norm \cite{hoyer2004nnm, lee2001ann} of
the factorization error.
\begin{equation}
\label{eq_L2_nmf}
\min_{W,H} ||V-WH||_2
\end{equation}
This is the objective function we use in the experiments  for this paper.

\subsection{Solving the optimization problem of NMF.}
Although in the current literature it is widely believed that NMF is
a non-convex problem and only local minima can be found, we will
show in the following subsections that a convex formulation does
exist. Despite the existence of the convex formulation, we also show
that a formulation of the problem as a generalized geometric program, which is non-convex,
could give a better approach for finding the global optimum.
\subsubsection{NMF as a convex conic program.}
\label{convex_nmf}
\begin{theorem}
The set of Completely Positive Matrices $\mathcal{K^{CP}}$ is a convex cone.
\end{theorem}
\begin{proof}
See \cite{berman2003cpm}p.71.
\end{proof}
It is always desirable to find the minimum rank of NMF since we are looking for the most compact representation of the data matrix $V$. Finding the minimum rank NMF can be cast as the following optimization problem:
\begin{eqnarray}
\min_{\mathcal{W,H}} \quad \mbox{rank}
        \left[ \begin{array}{cc}
          \mathcal{W} & V \\
          V^T & \mathcal{H}
        \end{array} \right]  & & \\
\nonumber \mbox{subject to:} & & \\
\nonumber & & \mathcal{W} \in \mathcal{K^{CP}} \\
\nonumber & & \mathcal{H} \in \mathcal{K^{CP}} \\
\end{eqnarray}
Since minimizing the rank is non-convex, we can use its convex
envelope that according to \cite{recht2007gmr} is the trace of the
matrix. So a convex relaxation of the above problem is:
\begin{eqnarray}
\min_{\mathcal{W,H}} \quad \mbox{Trace}(
        \left[ \begin{array}{cc}
          \mathcal{W} & V \\
          V^T & \mathcal{H}
        \end{array} \right]) & & \\
\mbox{subject to:} & & \\
\nonumber & & \mathcal{W} \in \mathcal{K^{CP}} \\
\nonumber & & \mathcal{H} \in \mathcal{K^{CP}} \\
\end{eqnarray}
After determining $\mathcal{W, H}$, $W$ and $H$ can be recovered by
CP factorization of $\mathcal{W, H}$, which again is not an easy
problem. In  fact there is no practical barrier function known yet
for the CP cone so that Interior Point Methods can be employed.
Finding a practical description of the CP cone is an open problem.
So although the problem is convex, there is no algorithm known for
solving it.

\subsection{Convex relaxations of the NMF  problem.}
In the following subsections we investigate convex relaxations of
the NMF problem with the Positive Semidefinite Cone
\cite{nemirovski}.
\subsubsection{A simple convex upper bound with Singular Value Decomposition.}
Singular Value Decomposition (SVD) can decompose a matrix in two factors $U, V$:
\begin{equation}
 A = U  V
\end{equation}
Unfortunately the sign of the SVD components of $A\geq 0$ cannot be
guaranteed to be non-negative except for the first eigenvector
\cite{minc1988nm}. However if we project $U, V$ on the nonnegative
orthant ($U,V\geq0$) we get a very good estimate (upper bound) for NMF. We will
call it clipped SVD,  (CSVD). CSVD was used as a benchmark for the
relaxations that follow. It has also been used as an initializer for
NMF algorithms \cite{langville2006inm}.

\subsubsection{Relaxation with a positive semidefinite cone.}
In the minimization problem of eq.~\ref{eq_L2_nmf} where the cost
function is the $L_2$ norm, the nonlinear terms $w_{il}h_{lj}$
appear. A typical way to get these terms \cite{nemirovski} would be
to generate a large vector $z=[W'(:); H(:)]$, where we use the
MATLAB notation ($H(:)$ is the column-wise unfolding of a matrix).
If $Z=zz^T$ ($rank(Z)=1$) and $z>0$ is true, then the terms appearing in
$||V-WH||_2$ are linear  in $Z$. In the following example
eq.~\ref{big_sdp1},~\ref{big_sdp2} (see next page) where $V \in \Re^{2 \times 3}, W
\in \Re^{2 \times 2}, H \in \Re^{2 \times 3}$ we show the structure
of $Z$. Terms in bold are the ones we need to express the constraint
$V=WH$, i.e $v_{11} = w_{11}h_{11}+w_{12}h_{21}$.
\begin{equation}
\label{big_sdp1}
  z=
  \left[
    \begin{array}{c}
      w_{11} \\
      w_{12} \\
      w_{21} \\
      w_{22} \\
      h_{11} \\
      h_{21} \\
      h_{12} \\
      h_{22} \\
      h_{13} \\
      h_{23}
    \end{array}
    \right]
    \end{equation}

\begin{figure*}[!ht]
\begin{equation}
\label{big_sdp2}
 \small
   Z=
   \left[
   \begin{array}{cccccccccc}
     w^2_{11} & w_{11}w_{12} & w_{11}w_{21} & w_{11}w_{22} & \mathbf{w_{11}h_{11}} & w_{11}h_{21} & \mathbf{w_{11}h_{12}} & w_{11}h_{22} & \mathbf{w_{11}h_{13}} & w_{11}h_{23} \\
     w_{12}w_{11} & w^2_{12} & w_{12}w_{21} & w_{12}w_{22} & w_{12}h_{11} & \mathbf{w_{12}h_{21}} & w_{12}h_{12} & \mathbf{w_{12}h_{22}} & w_{12}h_{13} & \mathbf{w_{12}h_{23}} \\
     w_{21}w_{11} & w_{21}w_{12} & w^2_{21} & w_{21}w_{22} & \mathbf{w_{21}h_{11}} & w_{21}h_{21} & \mathbf{w_{21}h_{12}} & w_{21}h_{22} & \mathbf{w_{21}h_{13}} & w_{21}h_{23} \\
     w_{22}w_{11} & w_{22}w_{12} & w_{22}w_{21} & w^2_{22} & w_{22}h_{11} & \mathbf{w_{22}h_{21}} & w_{22}h_{12} & \mathbf{w_{22}h_{22}} & w_{22}h_{13} & \mathbf{w_{22}h_{23}} \\
     h_{11}w_{11} & h_{11}w_{12} & h_{11}w_{21} & h_{11}w_{22} & h^2_{11} & h_{11}h_{21} & h_{11}h_{12} & h_{11}h_{22} & h_{11}h_{13} & h_{11}h_{23} \\
     h_{21}w_{11} & h_{21}w_{12} & h_{21}w_{21} & h_{21}w_{22} & h_{21}h_{11} & h^2_{21} & h_{21}h_{12} & h_{21}h_{22} & h_{21}h_{13} & h_{21}h_{23} \\
     h_{12}w_{11} & h_{12}w_{12} & h_{12}w_{21} & h_{12}w_{22} & h_{12}h_{11} & h_{12}h_{21} & h^2_{12} & h_{12}h_{22} & h_{12}h_{13} & h_{12}h_{23} \\
     h_{22}w_{11} & h_{22}w_{12} & h_{22}w_{21} & h_{22}w_{22} & h_{22}h_{11} & h_{22}h_{21} & h_{22}h_{12} & h^2_{22} & h_{22}h_{13} & h_{22}h_{23} \\
     h_{13}w_{11} & h_{13}w_{12} & h_{13}w_{21} & h_{13}w_{22} & h_{13}h_{11} & h_{13}h_{21} & h_{13}h_{12} & h_{13}h_{22} & h^2_{13} & h_{13}h_{23} \\
     h_{23}w_{11} & h_{23}w_{12} & h_{23}w_{21} & h_{23}w_{22} & h_{23}h_{11} & h_{23}h_{21} & h_{23}h_{12} & h_{23}h_{22} & h_{23}h_{13} & h^2_{23}
   \end{array}
   \right]
\end{equation}
\end{figure*}

Now the optimization problem eq.~\label{eq_L2_nmf} is equivalent to:
\begin{eqnarray}
\label{big_convex_relaxation}
\min &&\sum_{i=1,j=1}^{i=N, j=m}\sum_{l=1}^{k} \left(Z_{ik+l,Nk+jk+l}-V_{ij}\right)^2  \\
\nonumber && \mbox{subject to:}  \\
\nonumber &&\mbox{rank(Z)}=1
\end{eqnarray}

This is not a convex problem but it can be easily be relaxed to \cite{fazel2001rmh}:

\begin{eqnarray}
\label{big_sdp_problem_rank_1}
\min \mbox{Trace}(Z) & & \\
\nonumber \mbox{subject to:} & & \\
\nonumber A\bullet Z &=&V_{ij} \\
\nonumber Z &\succeq& 0 \\
\nonumber Z &\succeq& zz^T \\
\nonumber Z &\geq& 0
\end{eqnarray}
where $A$ is a matrix that selects the appropriate elements  from
$Z$. Here is an example for a matrix $A$ that selects the elements
of $Z$ that should sum to the $V_{13}$ element:
\begin{equation}
A_{13} = \left[
\begin{array}{cc}
  \mathbf{0} & \begin{array}{cccccc}
        0 & 0 & 0 & 0 & 1 & 0 \\
        0 & 0 & 0 & 0 & 0 & 1 \\
        0 & 0 & 0 & 0 & 0 & 0 \\
        0 & 0 & 0 & 0 & 0 & 0
      \end{array}  \\
 \mathbf{0} & \mathbf{0}
\end{array}
\right]
\end{equation}

In the second formulation (\ref{big_sdp_problem_rank_1}) we have
relaxed $Z=zz^T$ with $Z\succeq zz^T$. The objective function tries
to minimize the rank of the matrix, while the constraints try to
match the values of  the given matrix  $V$. After solving the
optimization problem the solution  can be found on the first
eigenvector of $Z$. The quality of the relaxation depends on the
ratio of the first eigenvalue to sum of the rest. The positivity of
$Z$ will guarantee that the first eigenvector will  have elements
with the same sign according to the Peron Frobenious Theorem
\cite{minc1988nm}. Ideally if the rest of the eigenvectors are
positive they can also be included. One of the problems of this
method is the complexity. $Z$ is $(N+m)k \times (N+m)k$ and there
are $\frac{((N+m)k)((N+m)k-1)}{2}$ non-negative constraints. Very
quickly the problem becomes unsolvable.

In practice the problem as posed in \ref{big_convex_relaxation}
always gives $W$ and $H$ matrices that are rank one. After testing
the method exhaustively  with random matrices $V$ that either had a
product $V=WH$ representation or not the solution was always  rank
one on both $W$ and $H$. This was always the case with any of the
convex formulations presented in this paper. This is because there
is a missing constraint that will let the energy of the dot products
spread among dimensions. This is something that should characterize
the spectrum of $H$.

The $H$ matrix is often interpreted as the basis vectors of the
factorization and $W$ as the matrix that has the coefficients. It is
widely known that in nature spectral analysis is giving spectrum
that decays either exponentially $e^{-\lambda f}$ or more slowly
$1/f^{\gamma}$. Depending on the problem we can try different
spectral functions. In our experiments we chose the exponential one.
Of course the decay parameter $\lambda$ is something that should be
set adhoc. We experimented with several values of $\lambda$, but we
couldn't come up with a systematic, heuristic and practical rule. In
some cases the reconstruction error was low but in some others not.
Another relaxation that was necessary for making the optimization
tractable was to reduce the the non-negativity constraints only on
the elements that are involved in the equality constraints.

\subsubsection{Approximating the SDP cone with smaller ones.}
A different way to deal with the computational complexity of SDP
(eq.~\ref{big_sdp_problem_rank_1}) is to approximate the big SDP
cone $(N+m)k \times (N+m)k$ with smaller ones. Let $W_i$ be the
$i_{th}$ row of  $W$ and $H_j$ the $j_{th}$ column of $H$. Now
$z_{ij}=[W_i(:)'; H_j(:)]$ ($2k$ dimensional vector) and
$Z_{ij}=z_{ij}z_{ij}^T$ ($2k \times 2k$ matrix), or
\begin{equation}
Z_{ij}=
\left[
\begin{array}{cc}
W_i^T W_i & W_i^TH_j \\
W_i^TH_j  & H_jH_j^T
       \end{array}
\right]
\end{equation}
or it is better to think it in the form:
\begin{equation}
\label{zeta_local} Z_{ij}= \left[
\begin{array}{cc}
  \mathcal{W}_i & \mathcal{Z}_{WH} \\
  \mathcal{Z}_{WH} & \mathcal{H}_j
\end{array}
\right]
\end{equation}
and once $\mathcal{W,H}$ are found then $W_i, H_j$ can be found from
SVD decomposition of $\mathcal{W,H}$ and the quality of the
relaxation will be judged upon the magnitude of the first eigenvalue
compared to the sum of the others. Now the optimization problem becomes:
\begin{eqnarray}
  \min  \sum_{i=1}^{N}\sum_{j=1}^{m} \mbox{Trace}(Z_{ij})&&  \\
\nonumber   Z_{ij} &\geq& 0 \\
\nonumber   Z_{ij} &\succeq & 0 \\
\nonumber   A_{ij} \bullet Z_{ij} &=& v_{ij},\quad \forall i,j
\end{eqnarray}
The above method has $Nm$ constraints. In terms of storage it needs
\begin{itemize}
  \item $(N+m)$ symmetric positive definite $k \times k$ matrices for every row/column of $W,H$, which is $\frac{(N+m)k(k+1)}{2}$
  \item $Nm$ symmetric positive definite $k \times k$ matrices for every $W_iH_j$ product, which is $\frac{(Nm)k(k+1)}{2}$
\end{itemize}
In total the storage complexity is $O((N+m+Nm)\frac{k(k+1)}{2})$ which
is significantly smaller by an order of magnitude from
$O(\frac{(N+m)k((N+m)k-1)}{2})$ which is the complexity of the previous
method.  There is also significant improvement in the computational
part. The SDP problem is solved with interior point methods
\cite{nemirovski} that require the inversion of a symmetric positive
definite matrix at some point. In the previous method that would
require $O((N+m)^3k^3)$ steps, while with this method we have to
invert $Nm$ $2k \times 2k$ matrices, that would cost $Nm(2k)^3$.
Because of their special structure the actual cost is $(Nm)k^3+
\max(N,m)k^3=(Nm+\max(N,m))k^3$.

We know that $\mathcal{W}_i, \mathcal{H}_j \succeq 0$. Since
$Z_{ij}$ is PSD and according to Schur's complement  on eq.~\ref{zeta_local}:
\begin{equation}
\label{schur}
\mathcal{H}_j-\mathcal{Z}_{WH}\mathcal{W}_i^{-1}\mathcal{Z}_{WH} \succeq 0
\end{equation}
So instead of inverting (\ref{zeta_local}) that would cost $8k^3$ we
can invert \ref{schur}. This formulation gives similar results with
the big SDP cone and most of the cases the results are comparable to
the CSVD.

\subsubsection{NMF as a convex multi-objective problem.}
A different approach would be to find a convex set  in which the
solution of the NMF lives and search for it over there. Assume that
we want to match  $V_{ij}=W_i H_j=\sum_{l=1}^{m}W_{il} H_{lj}$. In
this section we show that by controlling the ratio of the $L_2/L_1$
norms of $W,H$ it is possible to find the solution to NMF. Define
$W_{il}H_{lj}=V_{ij,l}$ and $\sum_{l=1}^{k}V_{ij,l}=V_{ij}$. We form
the following matrix that we require to be PSD:
\begin{equation}
\left[
\begin{array}{ccc}
  1      & W_{il} & H_{lj} \\
  W_{il} & t_{il}    & V_{ij,l} \\
  H_{lj} & V_{ij,l}    & t_{jl}
\end{array}
\right]
\succeq 0
\end{equation}
If we use the Schur complement we have:
\begin{equation}
\left[
\begin{array}{cc}
  t_{il}-W_{il}^2 & V_{ij,l} - W_{il}H_{lj} \\
  V_{ij,l} - W_{il}H_{lj} & t_{jl}-H_{lj}^2
\end{array}
\right]
\succeq 0
\end{equation}
An immediate consequence is that
\begin{eqnarray}
  t_{il} &\geq& W_{il}^2 \\
  t_{jl} &\geq& H_{ll}^2 \\
  (t_{il}-W_{il}^2)(t_{jl}-H_{lj}^2) &\geq& (V_{ij,l} - W_{il}H_{lj})^2
\end{eqnarray}
In the above inequality we see that the $L_2$ error
$\sum_{i=1}^{N}\sum_{j=1}^{m}\sum_{l=1}^{k}(V_{ij,l} -
W_{il}H_{lj})^2$ becomes zeros if $t_{il}=W_{il}^2, t_{jl}=H_{il}^2,
\quad \forall t_{il}, t_{jl}$. In general we want to minimize $t$
while maximizing $||W||^2$ and $||H||^2$. $L_2$ norm maximization is
not convex, but instead we can maximize $\sum W_{il}, \sum H_{lj}$
which are equal to the $L_1$ norms since everything is positive.
This can be cast as convex multi-objective problem \footnote{also
known as vector optimization} on the second order cone
\cite{boyd2004co}.
\begin{eqnarray}
\nonumber \min && \left[
\begin{array}{c}
  \sum_{i=1}^{i=N}\sum_{l=1}^{k} t_{il} + \sum_{j=1}^{j=m}\sum_{l=1}^{k} t_{lj} \\
  -\sum_{i=1}^{i=N}\sum_{l=1}^{k} W_{il} -  \sum_{j=1}^{j=m}\sum_{l=1}^{k} H_{lj}
\end{array}
        \right] \\
& & \mbox{subject to}: \\
\nonumber & & \left[
\begin{array}{cc}
  t_{il}-W_{il}^2 & V_{ij,l} - W_{il}H_{lj} \\
  V_{ij,l} - W_{il}H_{lj} & t_{jl}-H_{lj}^2
\end{array}
\right] \succeq 0
\end{eqnarray}
Unfortunately multi-objective optimization problems, even when they
are convex, they have local minima that are not global. An
interesting direction would be to test the robustness of existing
multi-objective algorithms on NMF.

\subsubsection{Local solution of the non-convex problem.}
In the previous sections we gave several convex formulations and
relaxations of the NMF problem that unfortunately are either
unsolvable or they give trivial rank one solutions that are not
useful at all.

In practice the non-convex formulation of  eq.~\ref{eq_L2_nmf} (classic
NMF objective) along with other like the KL distance between $V$ and
$WH$ are used in practice \cite{lee2001ann}.  All of them are
non-convex and several methods have been recommended, such as
alternating least squares, gradient decent or active set methods
\cite{kim:nnm}. In our experiments we used the L-BFGS method that
scales very well for large matrices.

\subsubsection{NMF as a Generalized Geometric Program and it's Global Optimum.}
\label{sec_geometric_formulation} The objective function
(eq.~\ref{eq_L2_nmf}) can be written in the following form:
\begin{equation}
||V-WH||_2=\sum_{i=1}^{N}\sum_{j=1}^{m}\sum_{l=1}^{k}\left(V_{ij}-W_{il}H_{lj}\right)^2
\end{equation}
The above function is twice differentiable so according to
\cite{horst1996god} the function can be cast as the difference of
convex (d.c.) functions. The problem can be solved with general
off-the-shelf global optimization algorithms. It can also be
formulated as a special case of dc programming, the generalized
geometric programming. With the following transformation
$W_{il}=e^{\tilde{w}_{il}}, H_{lj}=e^{\tilde{h}_{lj}}$ the objective
becomes:
\begin{eqnarray}
& & ||V-WH||_2=\sum_{i=1}^{N}\sum_{j=1}^{m}\sum_{l=1}^{k}\left(V_{ij}-e^{\tilde{w}_{il}+\tilde{h}_{lj}}\right)^2 \\
\nonumber & &=\sum_{i=1}^{N}\sum_{j=1}^{m}V_{ij}^2+
\sum_{i=1}^{N}\sum_{j=1}^{m}\left(\sum_{l=1}^{k}e^{\tilde{w}_{il}+\tilde{h}_{lj}}\right)^2 \\
\nonumber &
&-2\sum_{i=1}^{N}\sum_{j=1}^{m}V_{ij}\left(\sum_{l=1}^{k}e^{\tilde{w}_{il}+\tilde{h}_{lj}}\right)
\end{eqnarray}
The first term is constant and it can be ignored for the
optimization. The other two terms:
\begin{eqnarray}
f(\tilde{w}_{il},\tilde{h}_{lj}) &=& \sum_{i=1}^{N}\sum_{j=1}^{m}\left(\sum_{l=1}^{k}e^{\tilde{w}_{il}+\tilde{h}_{lj}}\right)^2 \\
g(\tilde{w}_{il},\tilde{h}_{lj})
&=&2\sum_{i=1}^{N}\sum_{j=1}^{m}V_{ij}\left(\sum_{l=1}^{k}e^{\tilde{w}_{il}+\tilde{h}_{lj}}\right)
\end{eqnarray}
are convex functions also known as the exponential form of
posynomials \footnote{Posynomial is a product of positive variables
exponentiated in any real number} \cite{boyd2004co}. For the global
solution of the problem
\begin{equation}
\min_{\tilde{W},\tilde{H}} \quad f(\tilde{W},\tilde{H})-g(\tilde{W},
\tilde{H})
\end{equation}
the algorithm proposed in \cite{floudas2000dgo} can be employed. The
above algorithm uses a branch and bound scheme that is impractical
for high dimensional optimization problems  as it requires too many
iterations to converge. It is worthwhile though to compare it with
thelocal non-convex NMF solver on a small matrix. We tried to do NMF
of order 2 on the following random matrix:
\[
\left[\begin{array}{ccc}
        0.45 & 0.434 & 0.35 \\
        0.70 & 0.64 & 0.43 \\
        0.22 & 0.01 & 0.3
      \end{array}
\right]
\]
After 10000 restarts of the local solver the best error we got was
$2.7\%$ while the global optimizer very quickly gave $0.015\%$
error, which is 2 orders of magnitude less than the local optimizer.

Another direction that is not investigated in this paper is the
recently developed algorithm for Difference Convex problems by Tao
\cite{tao1996dcf} that has been applied successfully to other data
mining applications such as Multidimensional Scaling.
\cite{an2001dpa}.

\section{Isometric Embedding}
\label{sec_Isometric_Embedding} The key concept in Manifold Learning
(ML)is to represent a dataset in a lower dimensional space by
preserving the local distances. The differences between methods
Isomap \cite{tenenbaum2000ggf}, Maximum Variance unfolding (MVU)
\cite{weinberger2006ind}, Laplacian EigenMaps \cite{belkin2003led}
and Diffusion Maps \cite{coifman2006dm} is how they treat distances
between points that are not in the local neighborhood. For example
IsoMap preserves exactly the geodesic distances, while Diffusion
Maps preserves distances that are based on the diffusion kernel.
Maximum Furthest Neighbor Unfolding (MFNU) \cite{vasiloglou2008ssm}
that is a variant of Maximum Variance Unfolding, preserves
local distance and it tries to maximize the distance between
furthest neighbors. In this section we are going to present the MFNU
method as it will be the basis for building isoNMF.

\subsection{Convex Maximum Furthest Neighbor Unfolding.}

Weinberger formulated the problem of isometric unfolding as a
Semidefinite Programming algorithm \cite{weinberger2006ind}\footnote{A similar approach to learning metrics is given in \cite{rosales2006lsm}}. In
\cite{kulis2007flr} Kulis  presented a non-convex formulation
of the problem that requires less memory than the Semidefinite one.
He also claimed that the non-convex formulation is scalable. The
non-convex formulation has the same global optimum with the
Semidefinite one as proven in \cite{burer2003npa}. In
\cite{vasiloglou2008ssm} Vasiloglou presented experiments where he
verified the scalability of this formulation. A variant of MVU the
Maximum Furthest Neighbor Unfolding (MFNU) was also presented in the
same paper. The latest formulation tends to be more robust and
scalable than MVU, this is why we will employ it as the basis of
isoNMF. Both methods can be cast as a semidefinite programming
problem \cite{vandenberghe1996sp}.

Given a set of data $X \in \Re^{N\times d}$, where $N$ is the number
of points and $d$ is the dimensionality, the dot product or Gram
matrix is defined as $G=X X^T $. The goal is to find a new Gram
matrix $K$ such that $rank(K)<rank(G)$ in other words $K=\hat{X}
\hat{X}^T$ where $\hat{X} \in \Re^{N\times d'}$ and $d'<d$.  Now the
dataset is represented by $\hat{X}$ which has fewer dimensions than
$X$. The requirement of isometric unfolding is that the euclidian
distances in the $\Re^{d'}$ for a given neighborhood around every
point have to be the same as in the $\Re^d$. This is expressed in:
\[
 K_{ii}+K_{jj}-K_{ij}-K_{ji}= G_{ii}+G_{jj}-G_{ij}-G_{ji}, \forall i, j \in I_i
\label{isometry}
\]
where $I_i$ is the  set of the indices of  the neighbors of the
$ith$ point. From all the $K$ matrices MFNU chooses the one that
maximizes the distances between furthest neighbor pairs. So the
algorithm is presented as an SDP:
\begin{eqnarray*}
   \max_{K} \;\sum_{i=1}^{N}B_{i} \bullet K &  &  \\
   \mbox{subject to}  & & \\
   A_{ij} \bullet K &=& d_{ij} \quad \forall j \in I_i  \\
    K &\succeq& 0
\end{eqnarray*}
where the $A \bullet X=\mbox{Trace}(AX^T)$ is the dot product
between matrices. $A_{ij}$ has the following form:
\begin{equation}
\label{A_structure} \left[
\begin{array}{cccccc}
  1 & 0 & \dots & -1 & \dots & 0 \\
  0 & \ddots & 0 & \dots & 0 & 0 \\
  \vdots & 0 & \ddots &  0 & \dots & 0 \\
  -1  & \dots & 0 & 1 & \dots & 0 \\
  \vdots & 0 & \dots & 0 & \dots & 0 \\
  0 & \dots & \dots & 0 &\dots & 0
\end{array}
\right]
\end{equation}
and
\begin{equation}
d_{ij}=G_{ii}+G_{jj}-G_{ij}-G_{ji}
\end{equation}
$B_i$ has the same structure of $A_{ij}$ and computes the distance
of the $i_{th}$ point with its furthest neighbor.
 The last condition is just a centering constraint for the
covariance matrix. The new dimensions $\hat{X}$ are the eigenvectors
of $K$. In general MFNU gives Gram matrices that have compact
spectrum, at least more compact than traditional linear Principal
Component Analysis (PCA). The method behaves equally well with MVU. Both MVU and MFNU are convex so they converge to the global optimum.
Unfortunately this method can handle datasets of no more than
hundreds of points because of its complexity. In the following
section a non-convex formulation of the problem that scales better
is presented.

\subsection{The Non Convex Maximum Furthest Neighbor Unfolding.}
\label{NCMVU}

By replacing the constraint $K\succeq 0$  \cite{burer2003npa} with
an explicit rank constraint $K=RR^T$ the problem becomes non-convex
and it is reformulated to
\begin{eqnarray}
\max & &  \sum_{i=1}^{N} B_{i} \bullet RR^T \\
\nonumber& & \mbox{subject to:} \\
\nonumber & & A_{ij} \bullet RR^T=d_{ij}
\end{eqnarray}
In \cite{burer2003npa}, Burer proved that the above formulation has
the same global minimum with the convex one.

The above problem can be solved with the augmented Lagrangian method
\cite{nocedal1999no}.
\begin{eqnarray}
\mathcal{L}&=& -\sum_{i=1}^{N} B_{i} \bullet RR^T \\
\nonumber & & -\sum_{i=1}^{N}\sum_{\forall j \in I_i}\lambda_{ij}(A_{ij}\bullet RR^T-d_ij) \\
\nonumber & &+\frac{\sigma}{2}\sum_{i=1}^{N}\sum_{\forall j \in
I_i}(A_{ij}\bullet RR^T-d_{ij})^2
\end{eqnarray}
Our goal is to minimize the Lagrangian that's why the objective
function is $-RR^T$ and not $RR^T$

The derivative of the augmented Lagrangian is:
\begin{eqnarray}
\label{derrivative} & & \frac{\partial \mathcal{L}}{\partial R} =
-2\sum_{i=1}^{N} B_{i} \bullet R \\
\nonumber & & -2\sum_{i=1}^{N}\sum_{\forall j \in I_i}
\lambda_{ij}A_{ij}R \\
\nonumber & & 2\sigma\sum_{i=1}^{N}\sum_{\forall j \in I_i}
(A_{ij}\bullet RR^T-d_ij)A_{ij}R
\end{eqnarray}
Gradient descent is a possible way to solve the minimization of the
Lagrangian, but it is rather slow. The Newton method is also
prohibitive. The Hessian of this problem is a sparse matrix although
the cost of the inversion might be high it is worth investigating.
In our experiments we used the limited memory BFGS  (L-BFGS) method
\cite{liu1989lmb, nocedal1999no} that is known to give a good rate
for convergence. MFNU in this non-convex formulation behaves much
better than MVU. In the experiments presented in
\cite{vasiloglou2008ssm}, MFNU tends to find more often the global
optimum, than MVU. The experiments also showed that the method
scales well up to 100K points.

\section{Isometric NMF.}
\label{sec_Isometric_NMF} NMF and MFNU are optimization problems.
The goal of isoNMF is to combine these optimization problems in one
optimization problem. MFNU has a convex and a non-convex
formulation, while for NMF only a non-convex formulation that can be
solved is known.
\subsection{Convex isoNMF.}
By using the theory presented in section \ref{convex_nmf} we can
cast isoNMF as a convex problem:
\begin{eqnarray}
\quad \max_{\tilde{W},\tilde{H}} & &  \sum_{i=1}^{N} B_{i} \bullet
Z \\
\nonumber & & \mbox{subject to:} \\
\nonumber & & A_{ij} \bullet \tilde{W}=d_{ij} \\
\nonumber & & Z = \left[ \begin{array}{cc}
                    \tilde{W} & V \\
                    V^T & \tilde{H}
                  \end{array} \right] \\
\nonumber & & Z \in \mathcal{K^{CP}} \\
\nonumber & & \tilde{W} \in \mathcal{K^{CP}} \\
\nonumber & & \tilde{H} \in \mathcal{K^{CP}}
\end{eqnarray}
Then $W, H$ can be found by the completely positive factorization of
$\tilde{W}=WW^T, \tilde{H}=HH^T$. Again this problem although it is
convex, there is no polynomial algorithm known for solving it.

\subsection{Non-convex formulation of isoNMF.}
The non convex isoNMF can be cast as the following problem:
\begin{eqnarray}
 \max & &  \sum_{i=1}^{N} B_{i} \bullet WW^T \\
\nonumber & & \mbox{subject to:} \\
\nonumber & & A_{ij} \bullet WW^T=d_{ij} \\
\nonumber & & WH=V \\
\nonumber & & W \geq 0 \\
\nonumber & & H \geq 0
\end{eqnarray}
The augmented lagrangian with quadratic penalty function is the following:
\begin{eqnarray}
\mathcal{L}& = & -\sum_{i=1}^{N} B_{i} \bullet WW^T \\
\nonumber & &-\sum_{i=1}^{N}\sum_{\forall j \in I_i} \lambda_{ij}(A_{ij}\bullet WW^T-d_ij) \\
\nonumber & & -\sum_{i=1}^{N}\sum_{j=1}^{m}\mu_{ij}\sum_{l=1}^{k}(W_{ik}H_{kj}-V_{ij}) \\
\nonumber & & +\frac{\sigma_1}{2}\sum_{i=1}^{N}\sum_{\forall j \in I_i}(A_{ij}\bullet WW^T-d_{ij})^2 \\
\nonumber & & +
\frac{\sigma_2}{2}\sum_{i=1}^{N}\sum_{j=1}^{m}\sum_{l=1}^{k}(W_{il}H_{lj}-V_{ij})^2
\end{eqnarray}
The non-negativity constraints are missing from the Lagrangian. This is because we can enforce them through the limited bound BFGS also known as L-BFGS-B.
The derivative of the augmented Lagrangian is:
\begin{eqnarray}
\label{derrivative}
& &\frac{\partial \mathcal{L}}{\partial W} = -2\sum_{i=1}^{N} B_{i}W\\
\nonumber & &-2\sum_{i=1}^{N}\sum_{\forall j \in I_i}\lambda_{ij}A_{ij}W \\
\nonumber & &-\sum_{i=1}^{N}\sum_{j=1}^{m}\mu_{ij}H \\
\nonumber & &+2\sigma_1\sum_{i=1}^{N}\sum_{\forall j \in I_i} (A_{ij}\bullet WW^T-d_ij)A_{ij}W \\
\nonumber &
&+2\sigma_2\sum_{i=1}^{N}\sum_{j=1}^{m}\sum_{l=1}^{k}(W_{il}H_{lj}-V_{ij})H
\end{eqnarray}

\begin{eqnarray}
\label{derrivative}
& & \frac{\partial \mathcal{L}}{\partial H} = -\sum_{i=1}^{N}\sum_{j=1}^{m}\mu_{ij}H \\
\nonumber &
&+2\sigma_2\sum_{i=1}^{N}\sum_{j=1}^{m}\sum_{l=1}^{k}(W_{il}H_{lj}-V_{ij})W
\end{eqnarray}

\subsection{Computing the local neighborhoods.}
\label{Allnn}
 As already discussed in previous section MFNU and
isoNMF require the computation of all-nearest and all-furthest
neighbors.  The all-nearest neighbor problem is a special case of a
more general class of problems called N-body problems
\cite{gray2001nbp}. In the following sections we give a sort
description of the nearest neighbor computation. The actual
algorithm is a four-way recursion. More details can be found in
\cite{gray2001nbp}.

\subsubsection{Kd-tree.}
The kd-tree is a hierarchical partitioning structure for fast
nearest neighbor search \cite{friedman1977afb}. Every node is
recursively partitioned in two nodes until the points contained are
less than a fixed number. This is a leaf. Nearest neighbor search is
based on a top down recursion until the query point finds the
closest leaf. When the recursion hits a leaf then it searches
locally for a candidate nearest neighbor. At this point we have an
upper bound for the nearest neighbor distance, meaning that the true
neighbor will be at most as far away as the candidate one. As the
recursion backtracks it eliminates (prunes) nodes that there are
further away than the candidate neighbor. Kd-trees provide on the
average nearest neighbor search in $O(\log N)$ time, although for
pathological cases the kd-tree performance can asymptotically have
linear complexity like the naive method.

\subsubsection{The Dual Tree Algorithm for nearest neighbor computation.}
In the single tree algorithm the reference points are ordered on a
kd-tree. Every nearest neighbor computation requires $O(\log(N))$
computations. Since there are $N$ query points  the total cost is
$O(N\log(N))$. The dual-tree algorithm \cite{gray2001nbp} orders the
query points on  a tree too. If the query set and the reference set
are the same then they can  share the same tree. Instead of querying
a single point at a time the dual-tree algorithm always queries a
group of points that live in the same node. So instead of doing the
top-down recursion individually for every point it does it for the
whole group at once. Moreover instead of computing distances between
points and nodes it computes distances between nodes. This is the
reason why most of the times the dual-tree algorithm can prune
larger portions of the tree than the single tree algorithm. The
complexity of the dual-tree algorithm is empirically $O(N)$. If the
dataset is pathological then the algorithm can be of quadratic
complexity too. The pseudo-code for the algorithm is described in
fig.~\ref{nn_dualtree_algorithm}.

\begin{figure}[!htb]
\begin{boxedminipage}[c]{\linewidth}
\footnotesize
\begin{verbatim}
recurse(q : KdTree, r : KdTree) {
  if (max_nearest_neighbor_distance_in_node(q)
          < distance(q, r) {
     /* prune */
  } else if (IsLeaf(q)==true and IsLeaf(r)==true) {
    /* search for every point in q  node  */
    /* its nearest neighbor in the r node */
    /* at leaves we must resort to        */
    /* exhaustive search O(n^2)           */
    /*update the maximum_neighbor_distance_in_node(q)*/
  } else if (IsLeaf(q)==false and IsLeaf(r)=true {
    /* choose the child that is closer to r */
    /* and recurse first                    */
    recurse(closest(r, q.left, q.right), r)
    recurse(furthest(r, q.left, q.right), r)
  } else if (IsLeaf(q)==true and IsLeaf(r)==false) {
    /* choose the child that is closer to q  */
    /*  and recurse first */
    recurse(q, closest(q, r.left, r.right))
    recurse(q, furthest(q, r.left, r.right))
  } else {
    recurse(q.left,closest(q.left, r.left, r.right));
    recurse(q.left,furthest(q.left, r.left, r.right));
    recurse(q.right,closest(q.right, r.left, r.right));
    recurse(q.right,furthest(q.right, r.left, r.right));
  }
}
\end{verbatim}
\end{boxedminipage}
\caption{Pseudo-code for the dual-tree all nearest neighbor
algorithm} \label{nn_dualtree_algorithm}
\end{figure}

\begin{figure}[!htb]
\begin{boxedminipage}[c]{\linewidth}
\footnotesize
\begin{verbatim}
recurse(q : KdTree, r : KdTree) {
  if (min_furthest_neighbor_distance_in_node(q)
        < distance(q, r) {
     /* prune */
  } else if (IsLeaf(q)==true and IsLeaf(r)==true) {
    /* search for every point in q node its
    /* furthest neighbor in the r node */
    /* at leaves we must resort to     */
    /* exhaustive search O(n^2) */
    /*update the minimum_furthest_distance_in_node(q)*/
  } else if (IsLeaf(q)==false and IsLeaf(r)=true {
    /*choose the child that is furthest to r */
    /* and recurse first */
    recurse(furthest(r, q.left, q.right), r)
    recurse(closest(r, q.left, q.right), r)
  } else if (IsLeaf(q)==true and IsLeaf(r)==false) {
    /* choose the child that is furthest to q  */
    /* and recurse first */
    recurse(q, furthest(q, r.left, r.right))
    recurse(q, closest(q, r.left, r.right))
  } else {
    recurse(q.left,furthest(q.left, r.left, r.right));
    recurse(q.left,closest(q.left, r.left, r.right));
    recurse(q.right,furthest(q.right, r.left, r.right));
    recurse(q.right,closest(q.right, r.left, r.right));
  }
}
\end{verbatim}
\end{boxedminipage}
\caption{Pseudo-code for the dual-tree all furthest neighbor
algorithm} \label{fn_dualtree_algorithm}
\end{figure}
\subsubsection{The Dual Tree Algorithm for all furthest neighbor algorithm.}
Computing the furthest neighbor with the naive computation is also
of quadratic complexity. The use of trees can speed up the
computations too. It turns out that furthest neighbor search for a
single query point is very similar to the nearest neighbor search
presented in the original paper of kd-tree \cite{friedman1977afb}.
The only difference is that in the top-down recursion the algorithm
always chooses the furthest node. Similarly in the bottom up
recursion we prune a node only if the maximum distance between the
point and the node is smaller than the current furthest distance.
The pseudo code is presented in fig.~\ref{fn_dualtree_algorithm}.

\section{Experimental Results}
\label{sec_Experimental_Results} 
\begin{figure}[h]
\centerline{(a)\includegraphics[height=4.0cm]{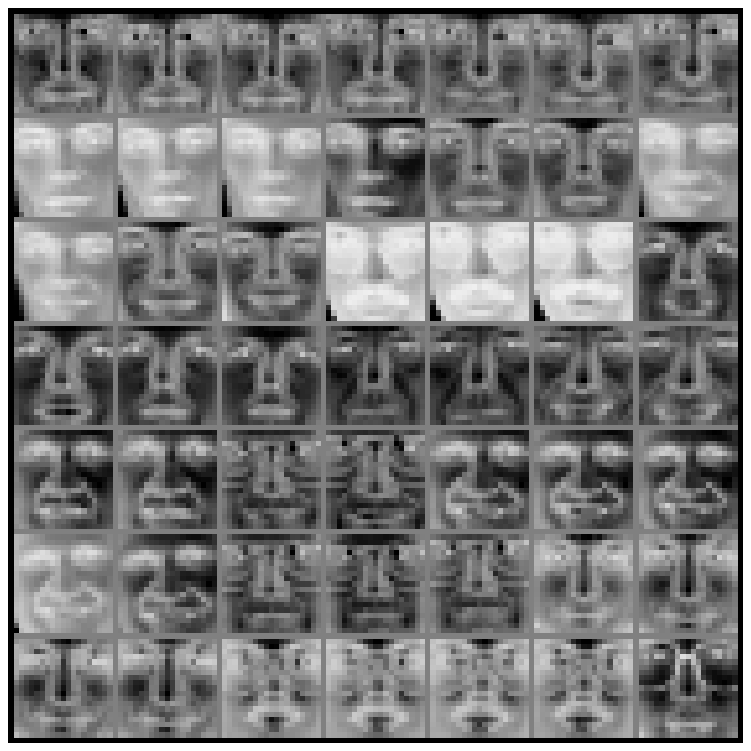} (b)\includegraphics[height=4.0cm]{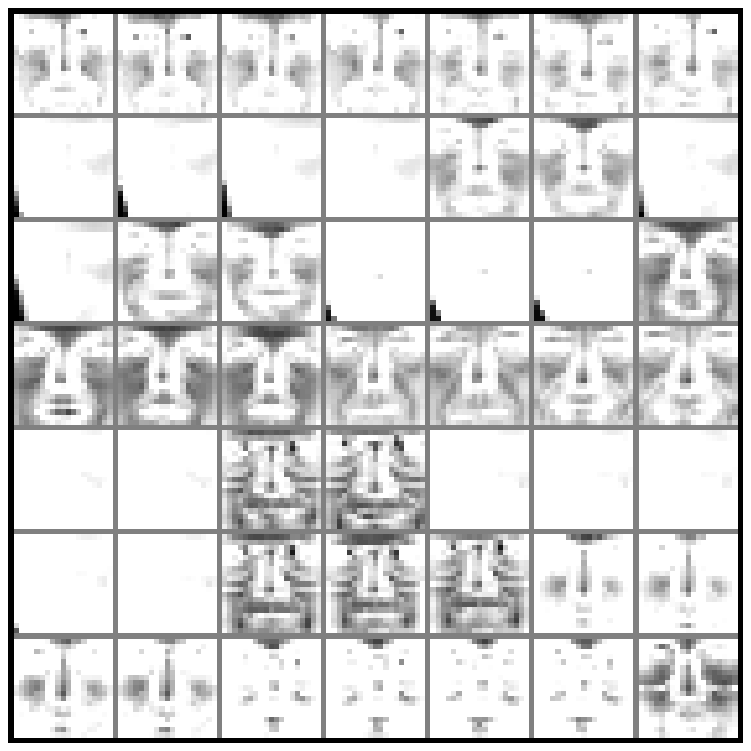}}
\centerline{(c)\includegraphics[height=4.0cm]{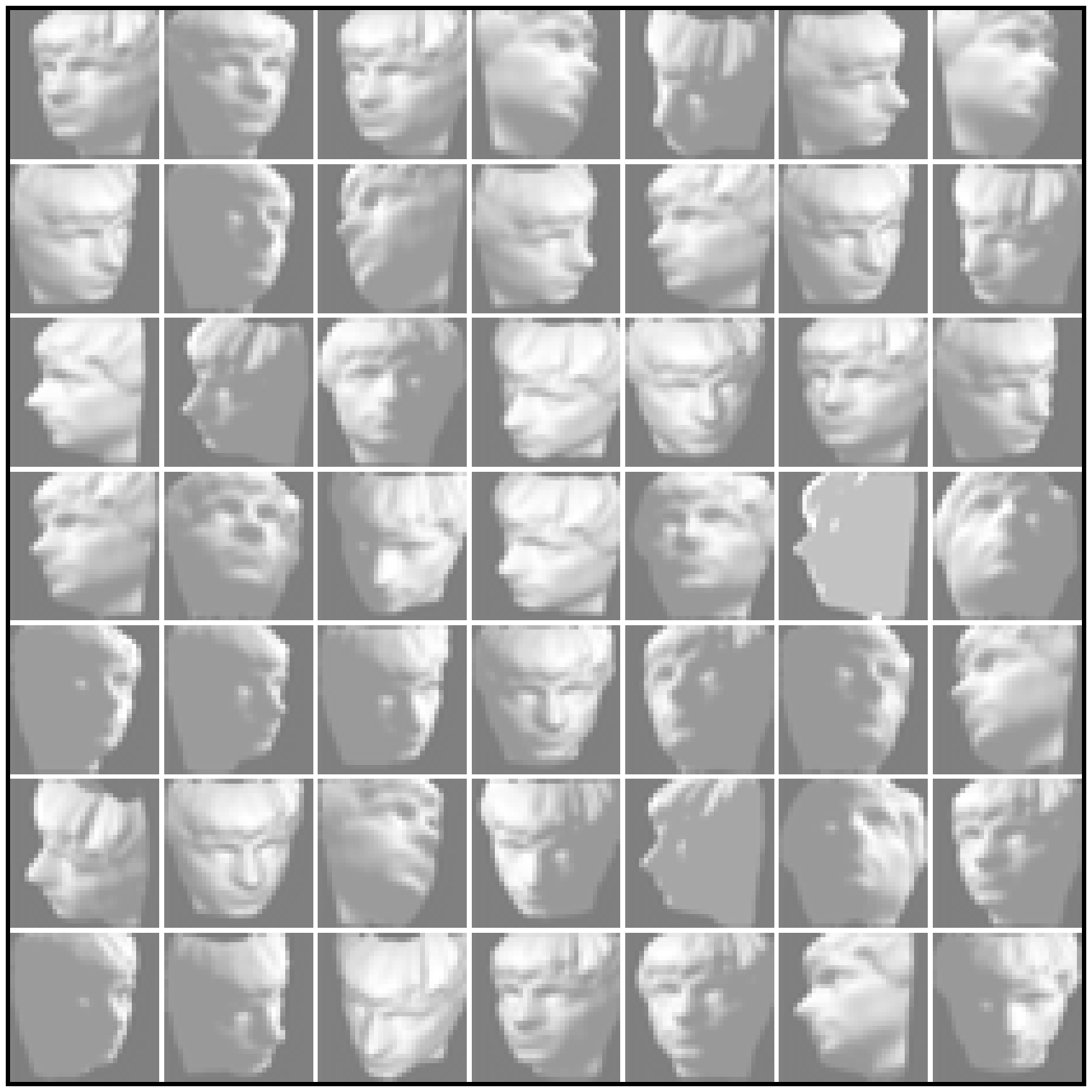} (d)\includegraphics[height=4.0cm]{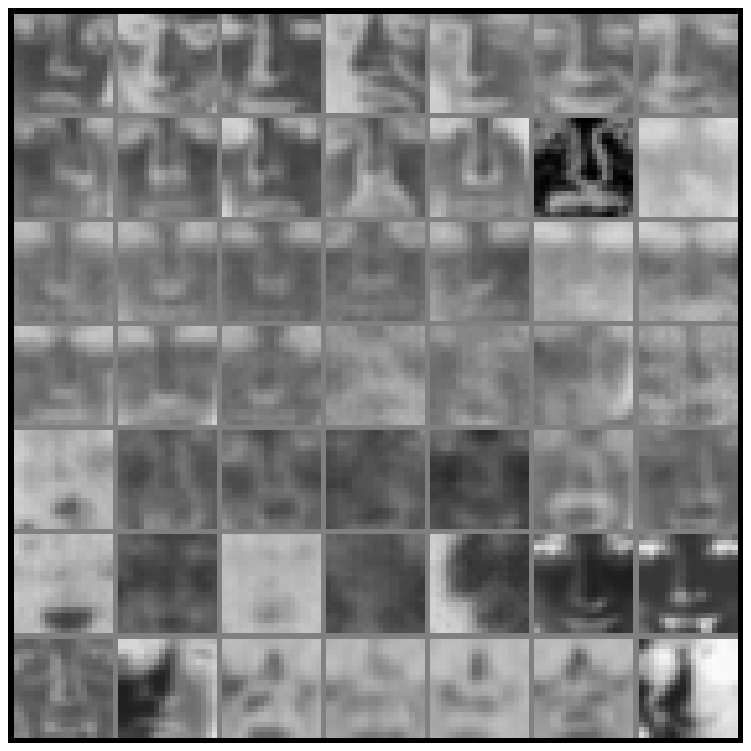}}
\caption{(a)Some images from the cbcl face database (b)The same images after variance normalization, mean set to 0.25 and thresholding in the interval [0,1] (c)The synthetic statue dataset from the isomap website \cite{isomap_statue} (d)472 images from the orl faces database \cite{orl_web}}
\label{raw_images}
\end{figure}

In order to evaluate and compare
the performance of isoNMF with traditional NMF we picked 3 benchmark
datasets that have been tested in the literature:
\begin{enumerate}
  \item The CBCL faces database fig.~\ref{raw_images}(a,b) \cite{cbcl_web},
used in the experiments of the original paper on NMF
\cite{lee1999lpo}. It consists of 2429 grayscale $19 \times 19$
images that they are hand aligned. The dataset was normalized as in
\cite{lee1999lpo}.
  \item The isomap statue dataset fig.~\ref{raw_images}(c)
\cite{isomap_statue} consists of 698  $64 \times 64$ synthetic face
photographed from different angles. The data was downsampled to $32
\times 32$ with the Matlab \emph{imresize} function (bicubic
interpolation).
  \item The ORL faces \cite{orl_web} fig.~\ref{raw_images}(d) presented in
\cite{hoyer2004nnm}. The set consists of  472 $19 \times 19$ gray
scale images that are not aligned. For visualization of the results
we used the \textit{nmfpack} code available on the web
\cite{nmfpack_web}.
\end{enumerate}

\begin{figure*}
\centerline{(a)\includegraphics[height=4.0cm]{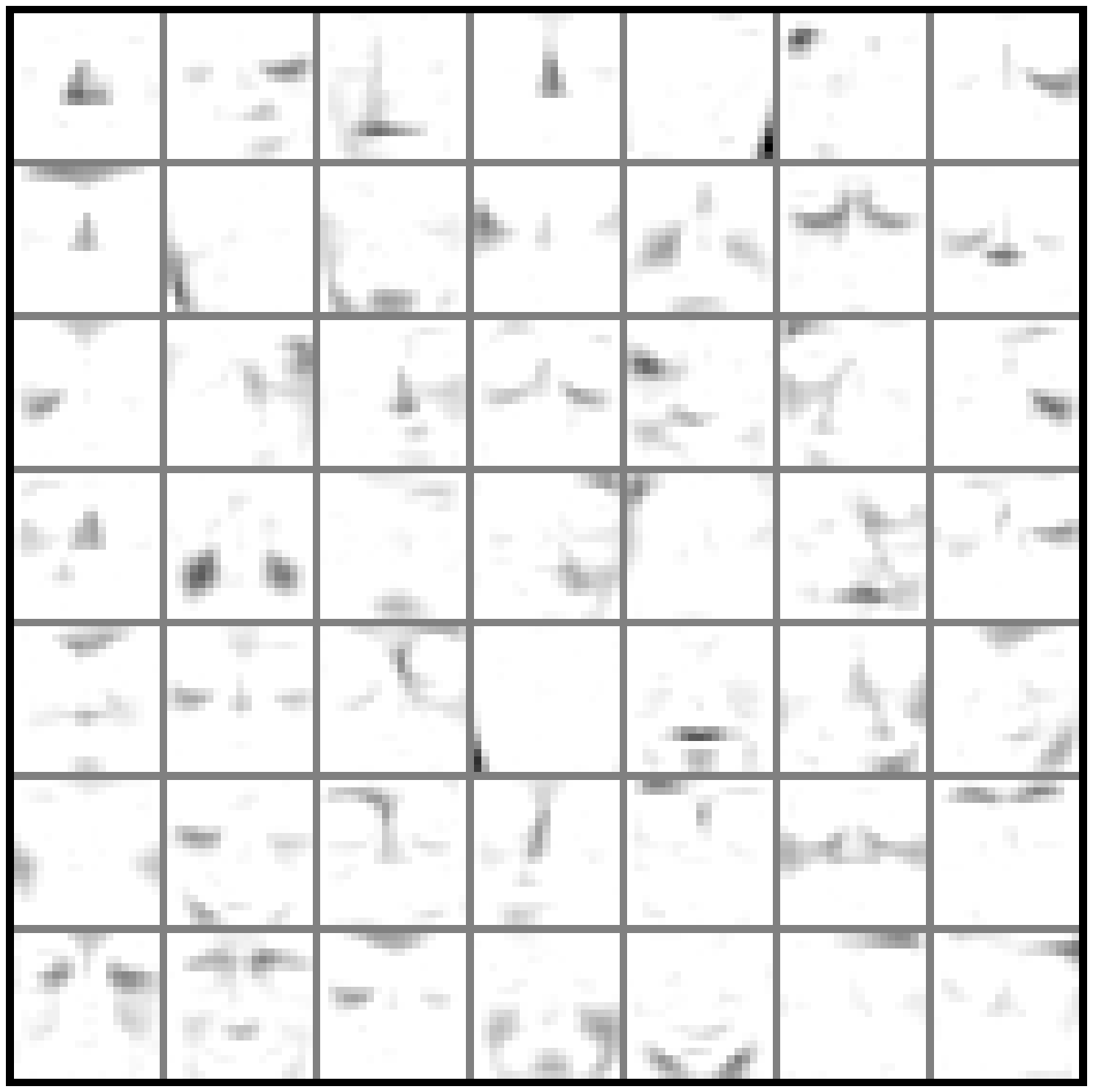} (b)\includegraphics[height=4.0cm]{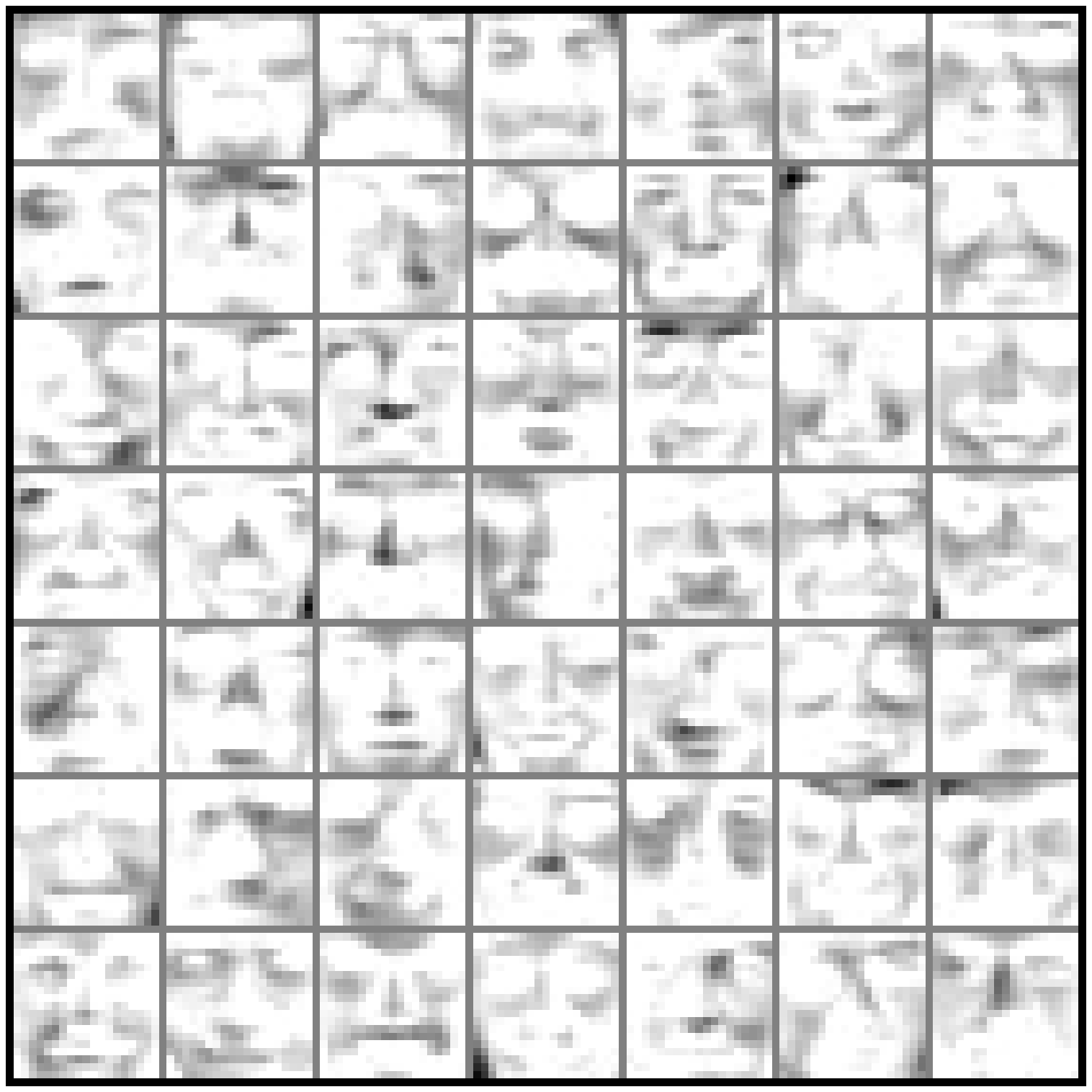} (c)\includegraphics[height=4.0cm]{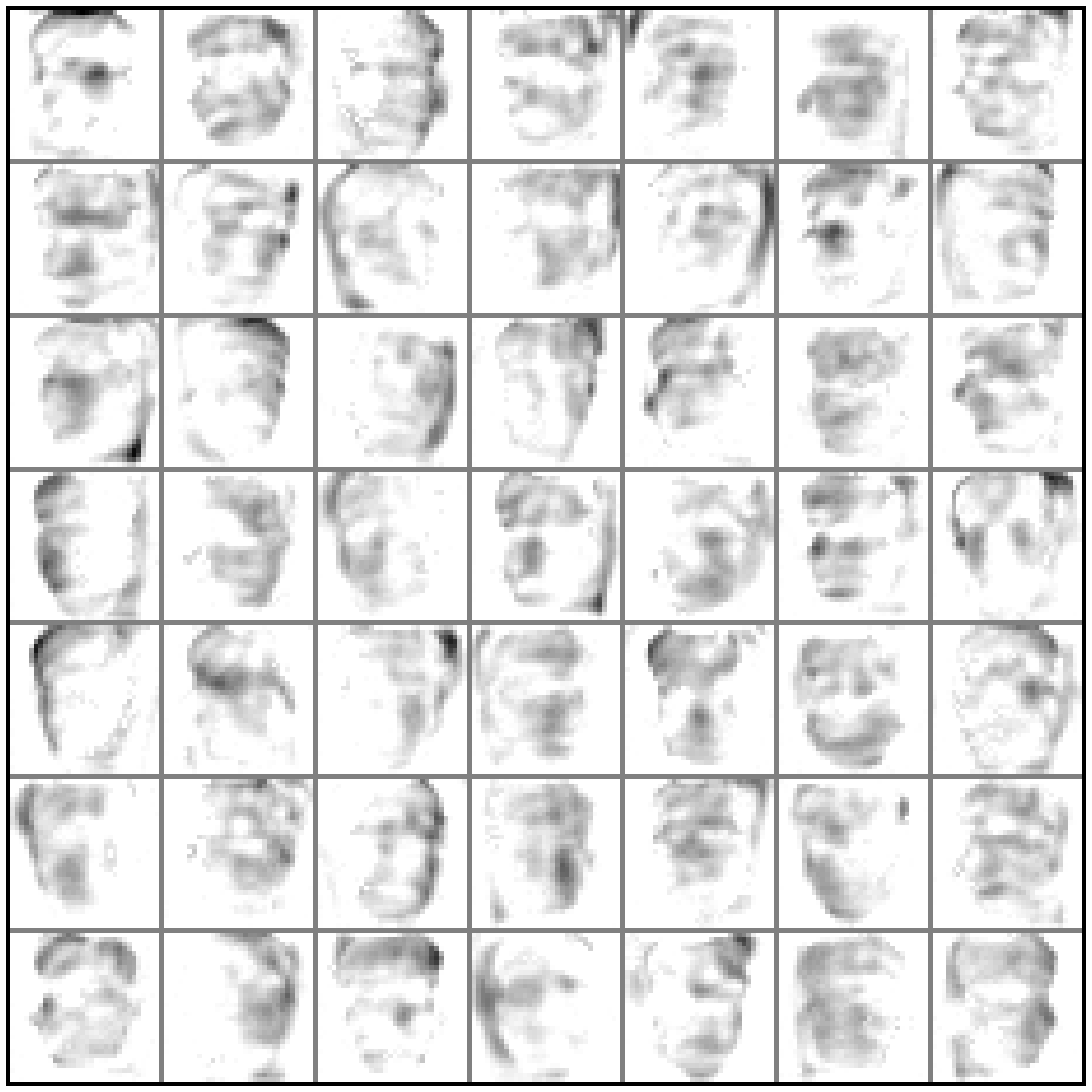} (d)\includegraphics[height=4.0cm]{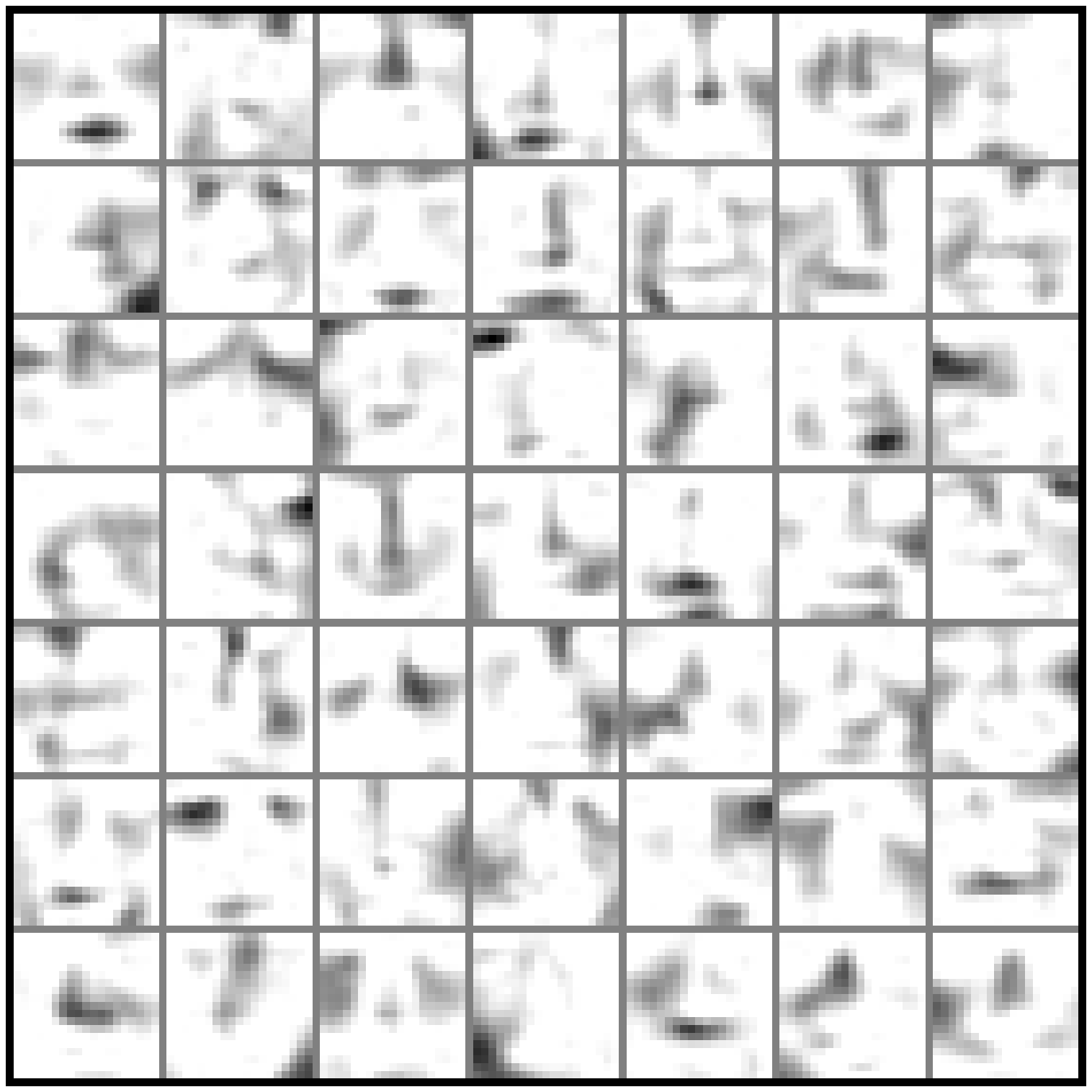}}
\centerline{(e)\includegraphics[height=4.0cm]{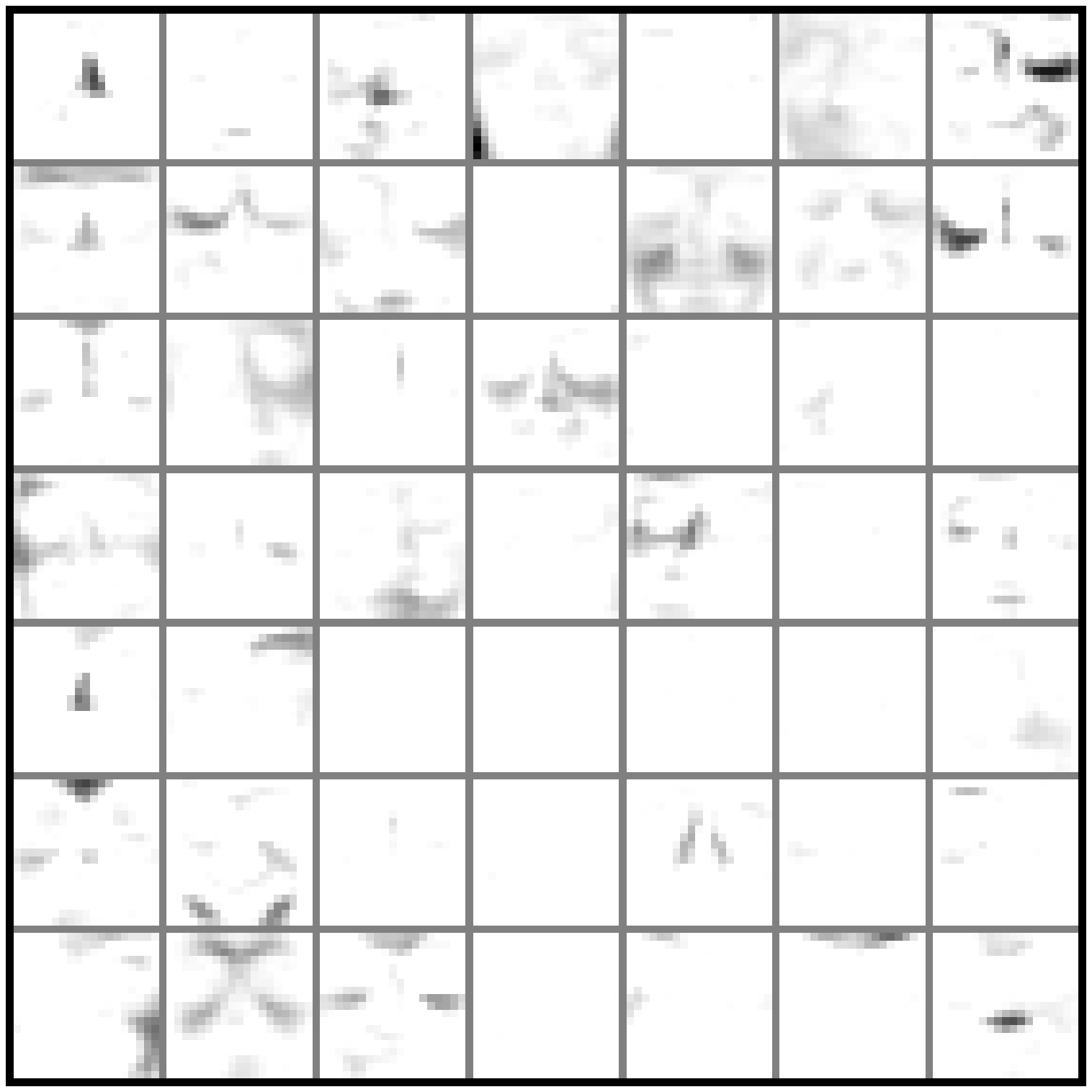} (f)\includegraphics[height=4.0cm]{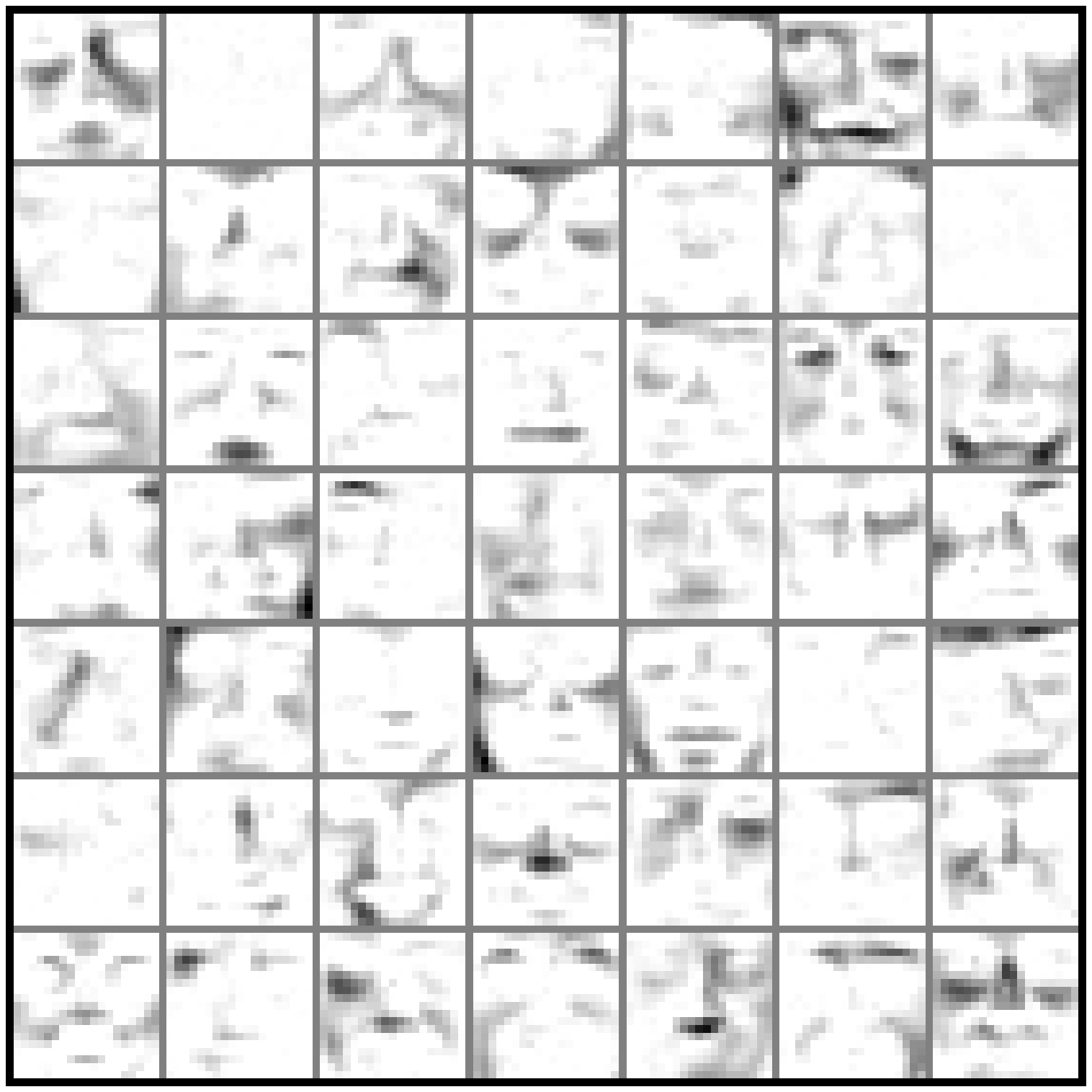} (g)\includegraphics[height=4.0cm]{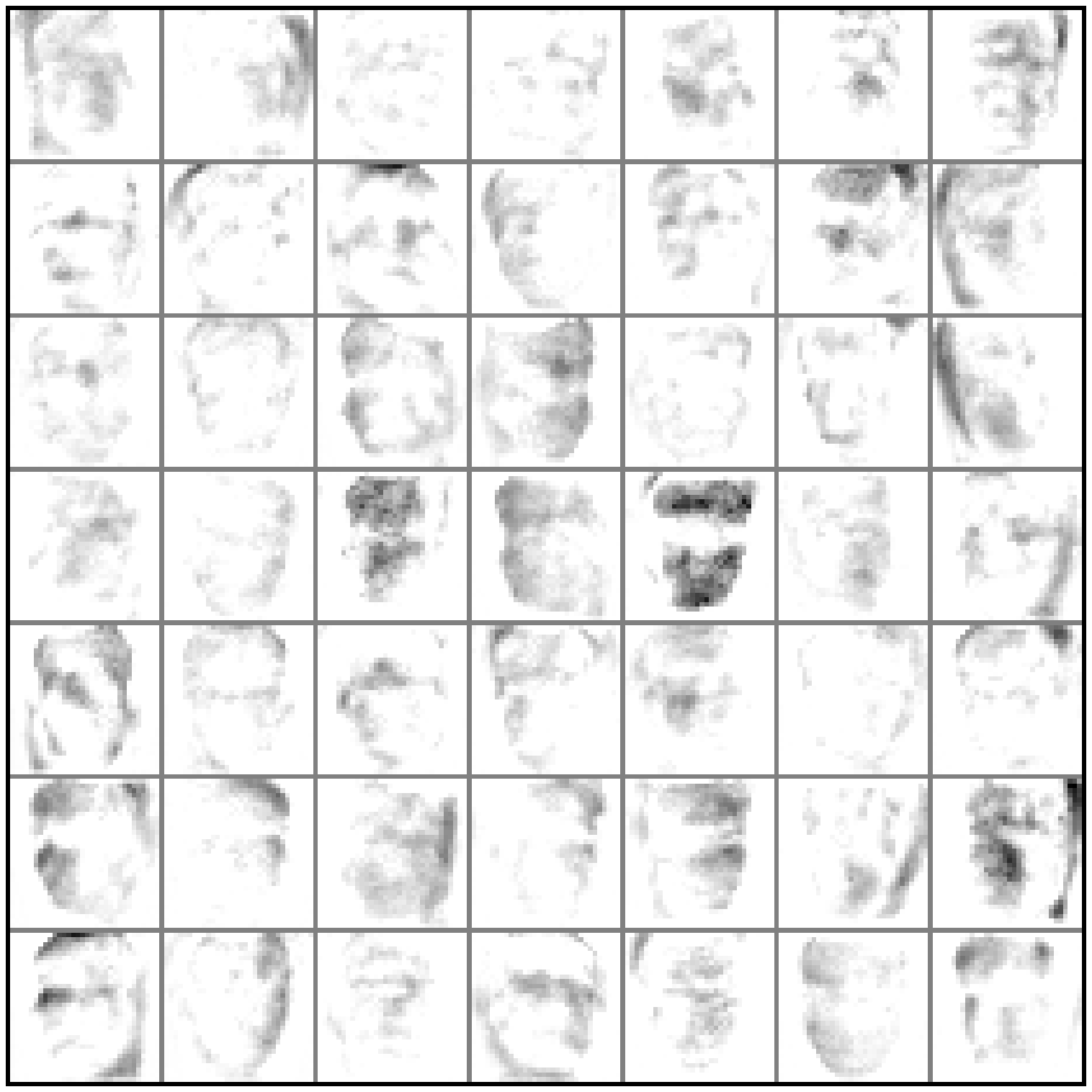} (h)\includegraphics[height=4.0cm]{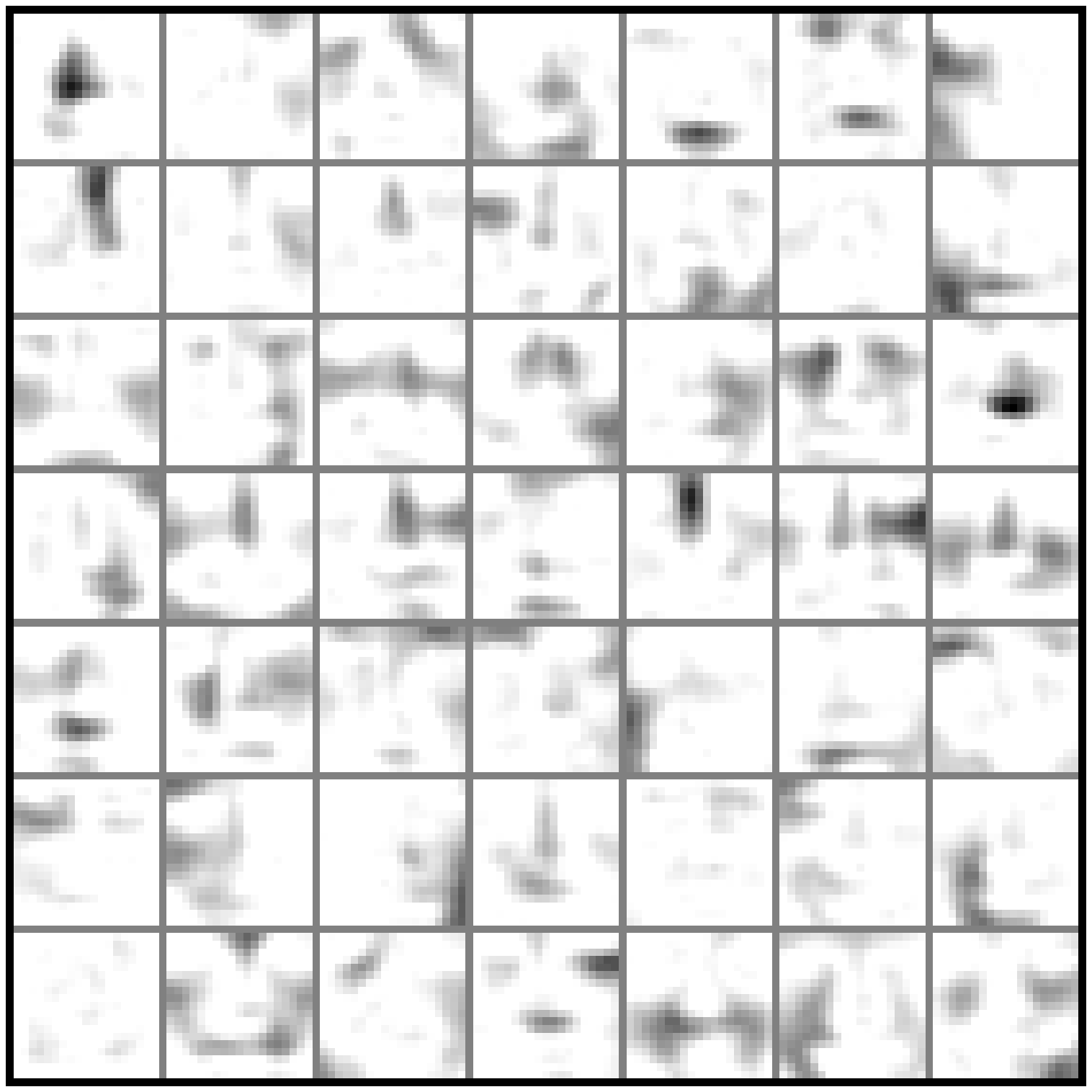}}
\caption{Top row: 49 Classic NMF prototype images. Bottom row: 49 isoNMF prototype images
$(a, e)$ CBCL-face database with mean variance normalization and thresholding, $(b, f)$ CBCL face database without preprocessing, $(c, g)$ Statue dataset $(d, h)$ORL face database}
\label{nmf_results}
\end{figure*}

The results for classic NMF and isoNMF with k-neighborhood equal to
3 are presented in fig.~\ref{nmf_results} and tables
\ref{nmf_metrics}, \ref{iso_nmf_metrics}. We observe that classic
NMF gives always lower reconstruction error rates that are not that
far away from the isoNMF. Classic NMF fails to preserve distances
contrary to isoNMF that always does a good job in preserving
distances. Another observation is that isoNMF gives more sparse
solution than classic NMF. The only case where NMF has a big
difference in reconstruction error is in the CBCL-face database when
it is being preprocessed. This is mainly because the preprocessing
distorts the images and spoils the manifold structure. If we don't
do the preprocessing fig.~\ref{nmf_results}(f), the reconstruction
error of NMF and isoNMF are almost the same. We would also like to point that isoNMF scales equally well with the classic NMF. Moreover they are seem to show the same sensitivity to the initial conditions.

\begin{figure}
\centerline{$(a)$\includegraphics[height=3.5cm]{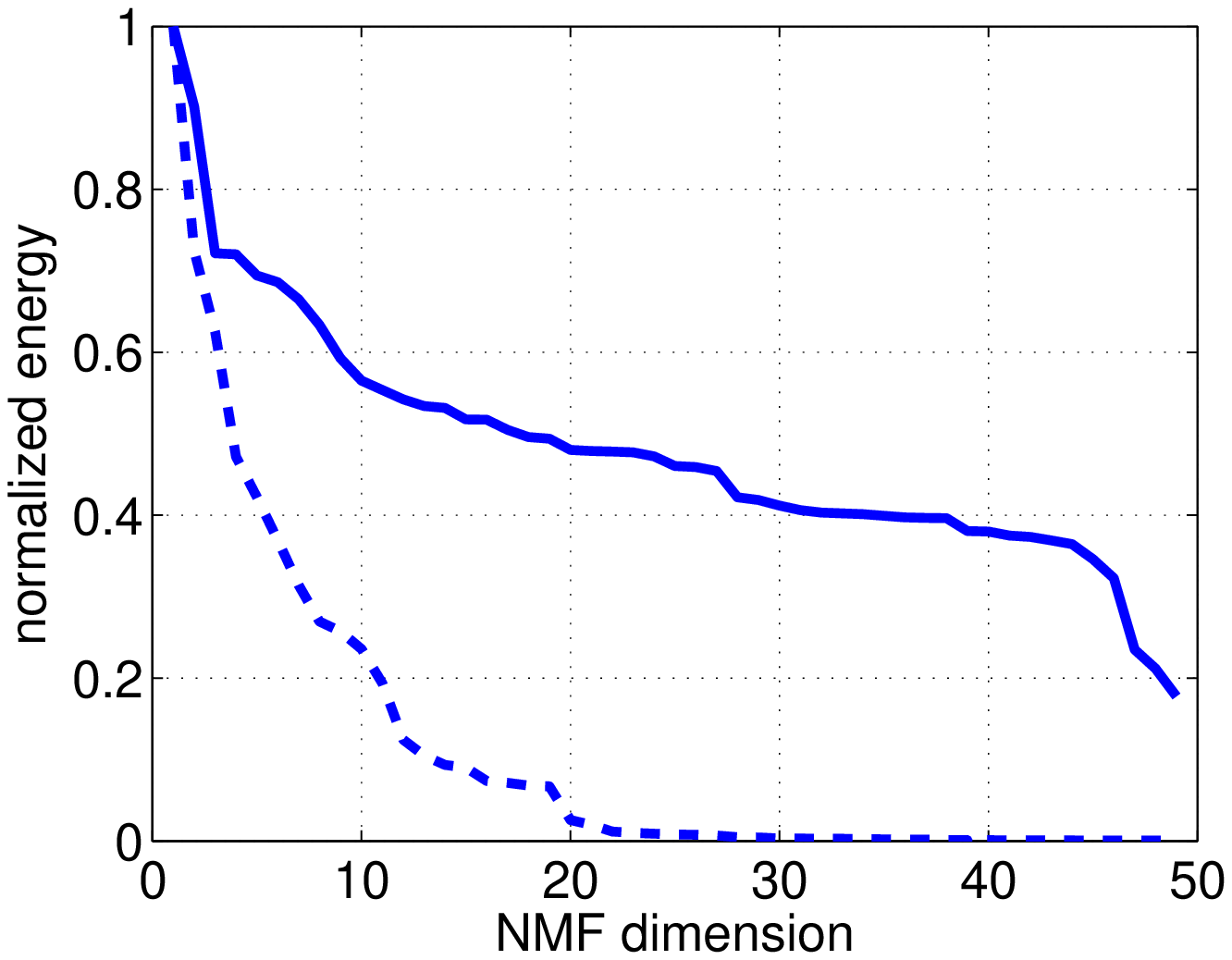} $(b)$\includegraphics[height=3.5cm]{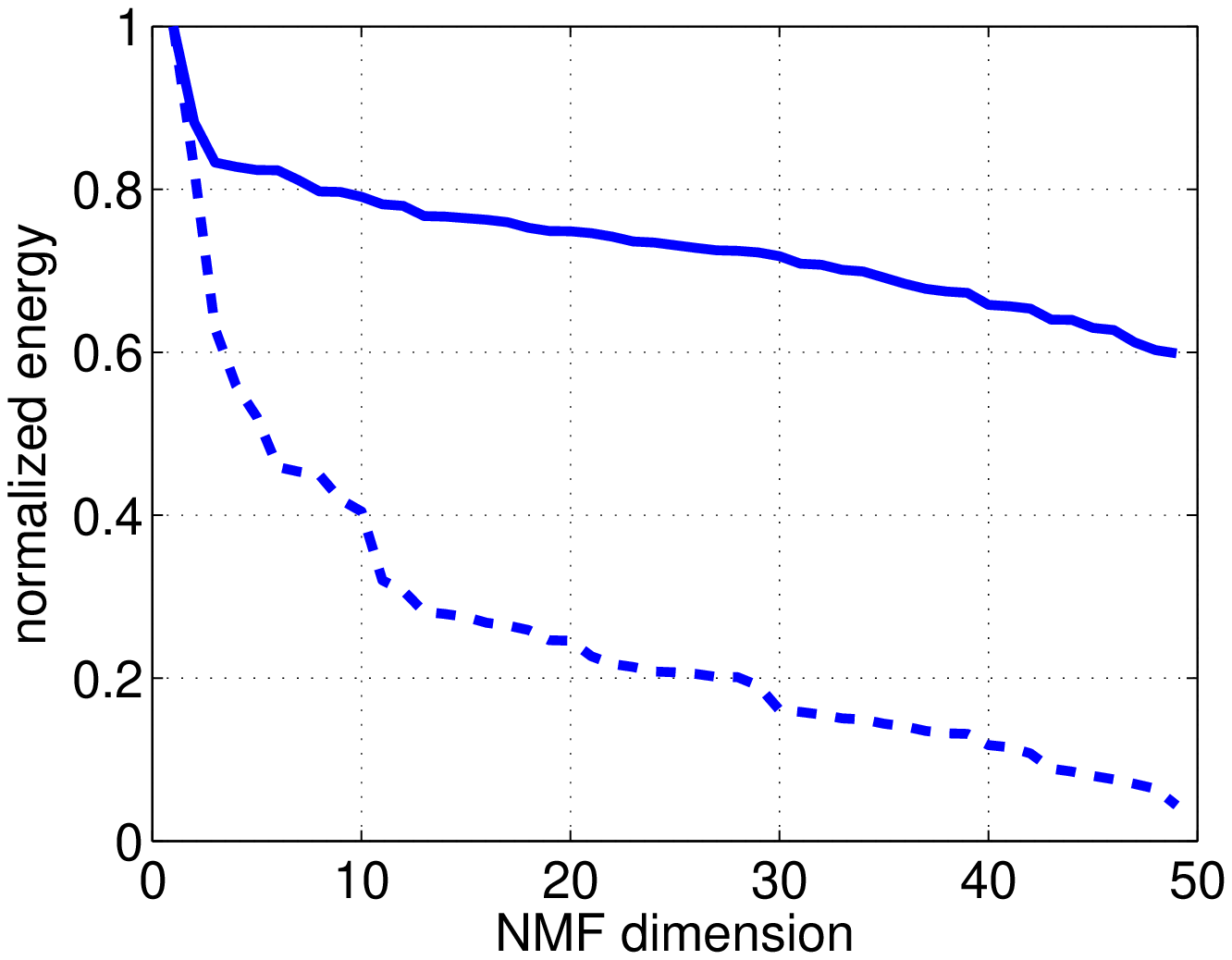}}
\centerline{$(c)$\includegraphics[height=3.5cm]{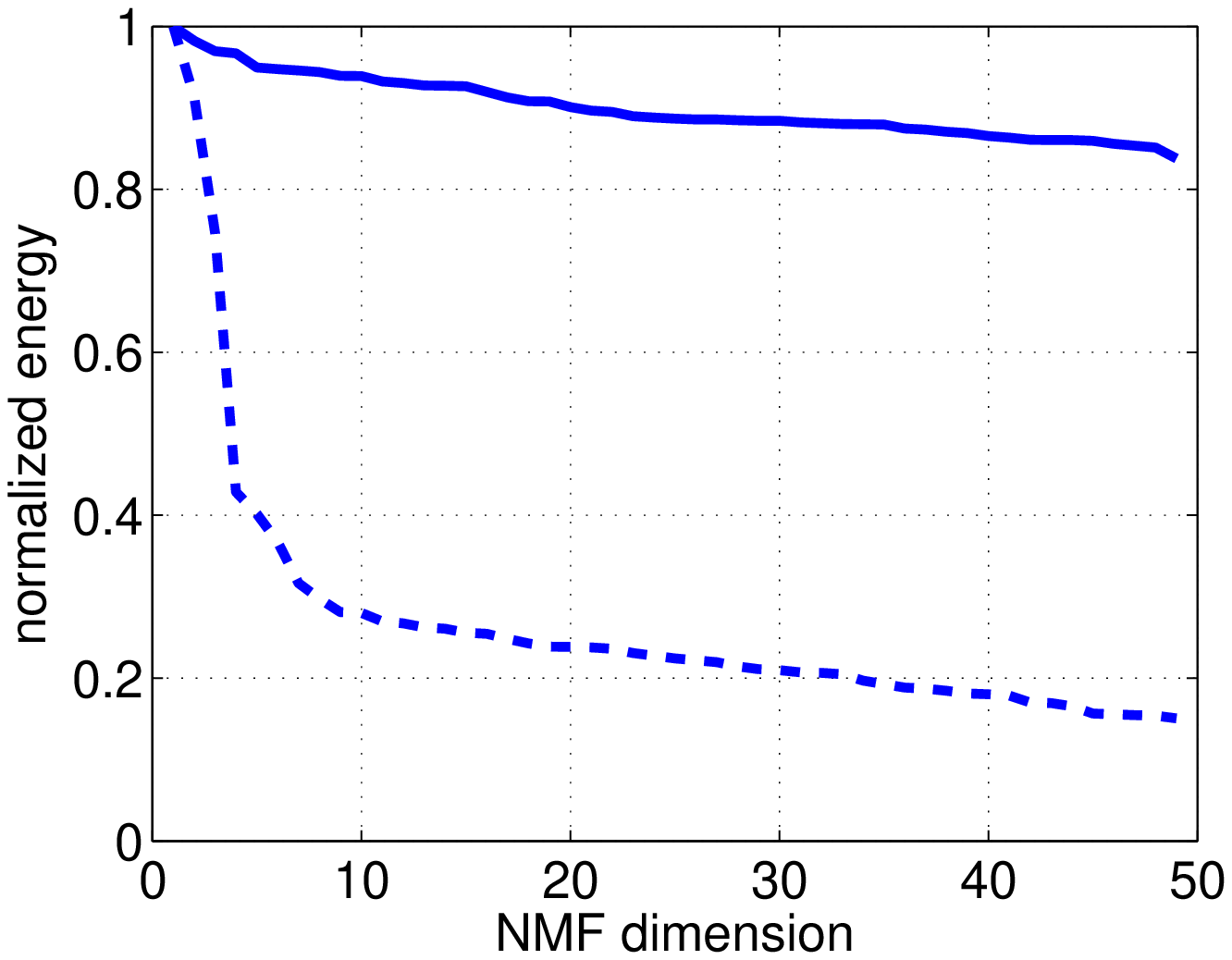}}
\caption{In this set of figures we show the spectrum of classic NMF (solid line) and Isometric NMF (dashed line) for the three datasets $(a)$cbcl face $(b)$isomap statue $(c)$orl faces. Although isoNMF gives much more compact spectrum we have to point that the basis functions are not orthogonal, so this figure is not comparable to SVD type spectrums}
\label{spectrums}
\end{figure}

In fig.~\ref{spectrums} we see a comparison of the energy spectrums
of classic NMF and isoNMF. We define the spectrum as
\[
  s_i=\frac{\sum_{l=1}^{N}W_{li}^2}{\sqrt{\sum_{l=1}^{M}H_{il}^2}}
\]
This represents the energy of the component normalized by the energy
of the prototype image generated by NMF/isoNMF. Although the results
show that isoNMF is much more compact than NMF, it is not a
reasonable metric. This is because the prototypes (rows of the $H$
matrix are not orthogonal to each other. So in reality
$\sum_{i=1}^{k}s_i < \sum_{i=1}^{N}\sum_{j=1}^{m}(WH)_{ij}^2$ and
actually much smaller. This is because the dot product between the
rows is not zero.

\begin{table}[!h]
\begin{tabular}{|c|c|c|c|c|}
  \hline
    classic NMF& cbcl norm. & cbcl    & statue  & orl      \\
  \hline
  rec. error           & 22.01\%    &  9.20\% & 13.62\% &  8.46\%  \\
  sparsity             & 63.23\%    & 29.06\% & 48.36\% & 46.80\%  \\
  dist. error          & 92.10\%    & 98.61\% & 97.30\% & 90.79\%  \\
  \hline
\end{tabular}
\caption{Classic NMF, the relative root mean square error, sparsity and distance error for the four different datasets (cbcl normalized and plain, statue and orl)}
\label{nmf_metrics}
\end{table}

\begin{table}[!h]
\begin{tabular}{|c|c|c|c|c|}
  \hline
    isoNMF             & cbcl norm. & cbcl    & statue  & orl    \\
  \hline
  rec error            & 33.34\%    & 10.16\% & 16.81\% & 11.77\% \\
  sparsity             & 77.69\%    & 43.98\% & 53.84\% & 54.86\% \\
  dist. error          & 4.19\%     &  3.07\% &  0.03\% &  0.01\% \\
  \hline
\end{tabular}
\caption{isoNMF, the relative root mean square error, sparsity and distance error for the four different datasets (cbcl normalized and plain, statue and orl)}
\label{iso_nmf_metrics}
\end{table}
%
%
%

\begin{figure}[t]
\centerline{(a)\includegraphics[height=4.0cm]{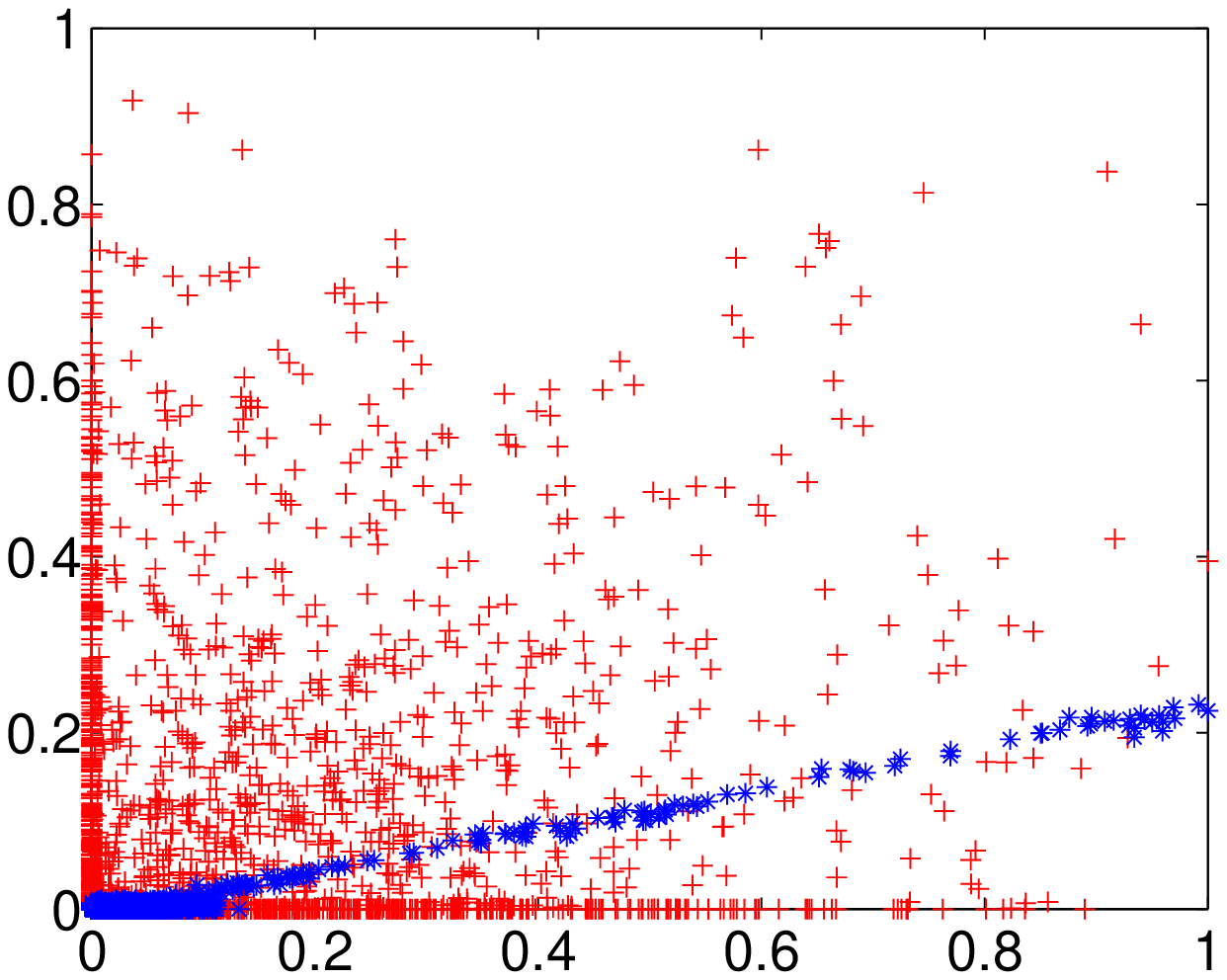}}
\centerline{(b)\includegraphics[height=4.0cm]{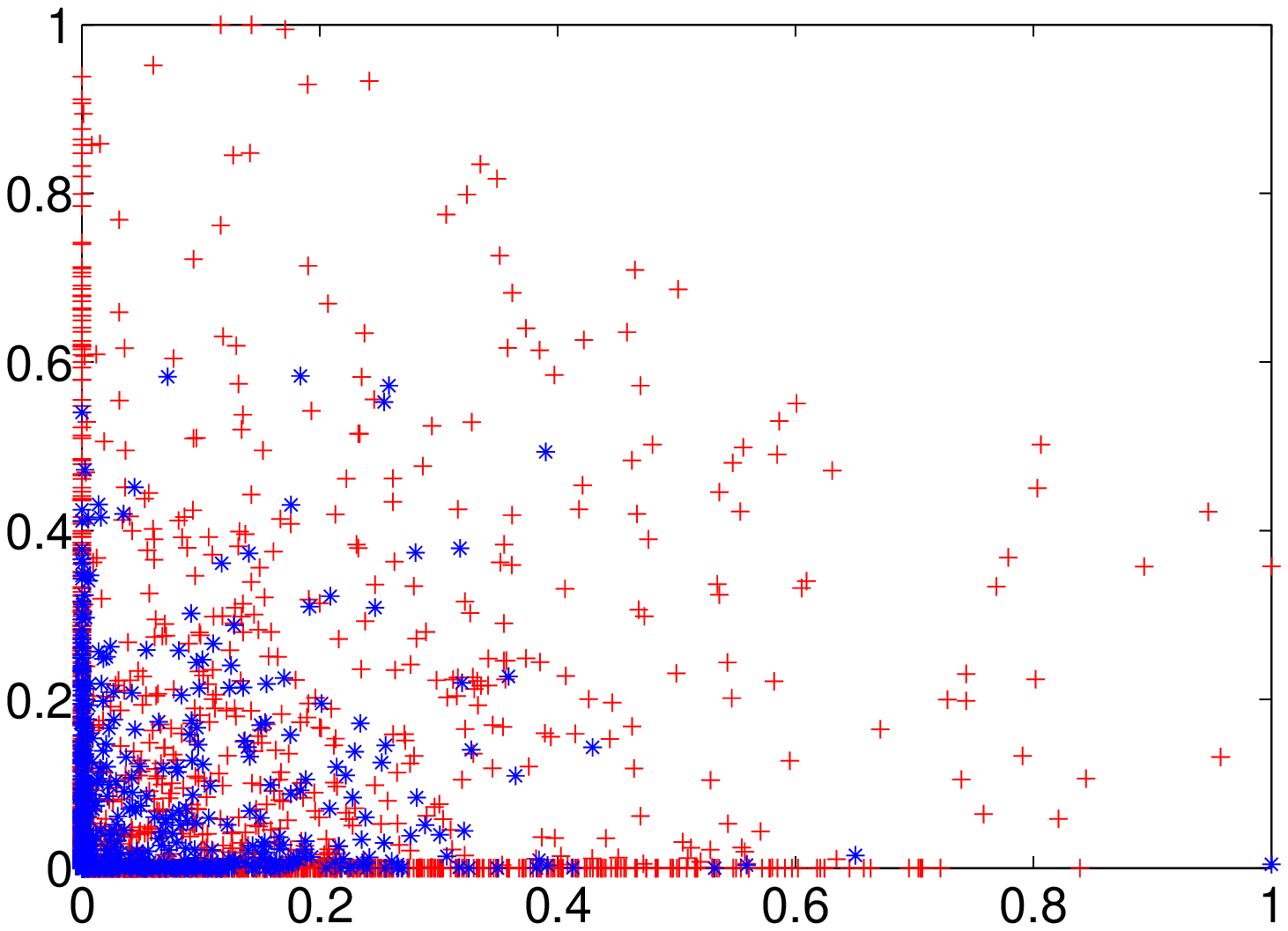}}
\centerline{(c)\includegraphics[height=4.0cm]{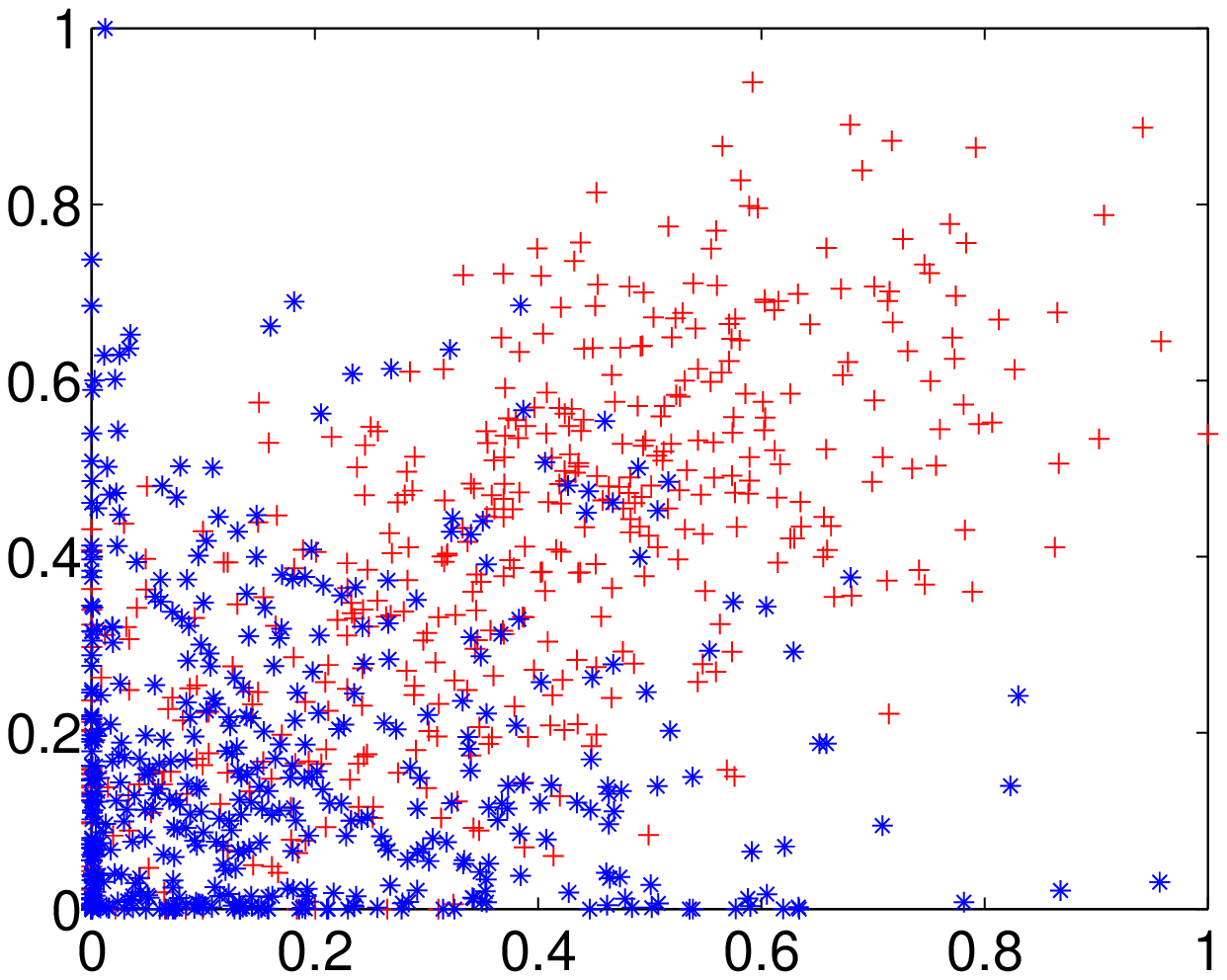}}
\caption{Scatter plots of two largest components of classic NMF(in blue) and Isometric NMF(in red) for $(a)$cbcl faces $(b)$isomap faces $(c)$orl faces}
\label{scatter}
\end{figure}

\section{Conclusion}
In this paper we presented a deep study of the optimization problem
of NMF, showing some fundamental existence theorems for the first
time as well as various advanced optimization approaches -- convex
and non-convex, global and local.  We believe that this study has
the capability to open doors for further advances in NMF-like
methods as well as other machine learning problems.  We also
developed and experimentally demonstrated a new method, isoNMF,
which preserves both non-negativity and isometry, simultaneously
providing two types of interpretability.  With the added reliability
and scalability stemming from an effective optimization algorithm,
we believe that this method represents a potentially valuable
practical new tool for the exploratory analysis of common data such
as images, text, and spectra.
\bibliographystyle{plain}
\bibliography{main}
\end{document}